\documentclass{article}
\usepackage{arxiv}

\usepackage[utf8]{inputenc} 
\usepackage[T1]{fontenc}    
\usepackage{hyperref}       
\usepackage{url}            
\usepackage{booktabs}       
\usepackage{amsfonts}       
\usepackage{nicefrac}       
\usepackage{microtype}      
\usepackage{graphicx}
\usepackage{natbib}
\usepackage{doi}
\usepackage{caption}
\usepackage{graphicx}
\usepackage{amsmath,amssymb,amsfonts, bm}
\usepackage{algorithmic}
\usepackage{algorithm}

\title{Adaptive Nesterov Accelerated Distributional Deep Hedging for Efficient Volatility Risk Management}

\author{ {Lei Zhao}\\
	Department of Electrical and Computer Engineering\\
	University of Victoria\\
	\texttt{leizhao@uvic.ca} \\
	\And
	{Lin Cai} \\
	Department of Electrical and Computer Engineering\\
	University of Victoria\\
	\texttt{cai@ece.uvic.ca} \\
	\And
	{Wu-Sheng Lu} \\
	Department of Electrical and Computer Engineering\\
	University of Victoria\\
	\texttt{wslu@ece.uvic.ca} \\
}

\date{}

\begin{document}
\maketitle

\begin{abstract}
In the field of financial derivatives trading, managing volatility risk is crucial for protecting investment portfolios from market changes. Traditional Vega hedging strategies, which often rely on basic and rule-based models, are hard to adapt well to rapidly changing market conditions. We introduce a new framework for dynamic Vega hedging, the Adaptive Nesterov Accelerated Distributional Deep Hedging (ANADDH), which combines distributional reinforcement learning with a tailored design based on adaptive Nesterov acceleration. This approach improves the learning process in complex financial environments by modeling the hedging efficiency distribution, providing a more accurate and responsive hedging strategy. The design of adaptive Nesterov acceleration refines gradient momentum adjustments, significantly enhancing the stability and speed of convergence of the model. Through empirical analysis and comparisons, our method demonstrates substantial performance gains over existing hedging techniques. Our results confirm that this innovative combination of distributional reinforcement learning with  the proposed optimization techniques  improves financial risk management and highlights the practical benefits of implementing advanced neural network architectures in the finance sector. 
\end{abstract}


\section{Introduction}
In modern financial markets, effective risk management is paramount for maintaining the stability and performance of investment portfolios~\cite{deng2016deep}\cite{hull2012risk}. Volatility risk, primarily quantified by implied volatility, plays an important role in the pricing and performance of financial instruments, especially options contracts~\cite{liu2019pricing}\cite{cao2023gamma}. These contracts provide mechanisms for traders to buy or sell an asset at a predetermined price within a specified timeframe. Due to the dynamic nature of  markets, the value of options is highly sensitive to changes in volatility, demanding the development of adaptive hedging strategies that can effectively manage risk~\cite{park2022uncertainty}. As financial models evolve, the incorporation of advanced neural network architectures and learning systems becomes crucial in designing strategies that not only predict but also mitigate the adverse effects of volatility fluctuation~\cite{andersen2017short}.

Traditional delta hedging primarily focuses on adjusting positions in the underlying asset to counteract changes in the option's value resulting from movements in the asset's price~\cite{alexander2023delta}. Although this method effectively manages the risk associated with price sensitivity, it often overlooks option price changes due to implied volatility in suboptimal risk management~\cite{rebonato2005volatility}. As a response, the financial sector is evolving towards more sophisticated hedging strategies that explicitly address volatility risk, underlining the importance of volatility management in safeguarding portfolios against market downturns~\cite{brownlees2010comparison}.

Vega hedging represents an advanced response to this challenge by targeting the sensitivity of option prices to shifts in implied volatility~\cite{herrmann2017model}. Traders adjust option positions to manage changes in Vega, i.e., the derivative of the option price with respect to changes in implied volatility, aiming to protect their portfolios from the adverse effects of volatility fluctuations~\cite{brunetti2016speculators}. Nonetheless, implementing Vega hedging encounters several challenges, including the need for accurate forecasts of future volatility and the necessity for dynamic position adjustments as market conditions evolve~\cite{figlewski1997forecasting}. Traditional Vega hedging methods, often reliant on heuristic-based approaches or the Black-Scholes model, are not effective to handle the complex dynamics of the market~\cite{klaas2019machine}.

Furthermore, traditional reinforcement learning (RL) approaches often fall short due to the need for precise risk management and adaptability in volatile financial environments~\cite{zhao2019routing}\cite{wang2019smart}. Hedging strategies require not only an optimized response to market shifts but also a deep understanding of the risk profile associated with potential outcomes. Distributional RL addresses these needs by modeling the full distribution of rewards rather than merely focusing on the expected value, providing richer insights into the variability and tail risks in hedging strategies~\cite{dabney2018distributional}. This capability is essential in financial applications where even minor deviations can lead to significant financial impacts. By capturing a range of possible outcomes and their associated probabilities, Distributional RL enables more robust decision-making under uncertainty, allowing for a nuanced response to changes in market volatility. This approach is particularly valuable in hedging scenarios, as it enhances an agent's ability to dynamically adapt while balancing risk and return more effectively than standard RL methods.

We propose an Adaptive Nesterov Accelerated Distributional Deep Hedging (ANADDH), a novel approach for dynamic Vega hedging that significantly improves the efficiency and stability of hedging strategies in financial derivatives trading. This innovative integration improves the state-of-the-art learning-based methods by incorporating additional momentum feedback, significantly enhancing stability and efficiency for real-time financial market applications. Adaptive Nesterov acceleration, a key element of our framework, ensures fast convergence and consistent model updates, enabling our strategies to quickly adjust to market volatilities and maintain optimal performance with lower computational demands. Our thorough empirical analyses show that our approach not only outperforms conventional strategies in managing volatility risks within financial derivatives portfolios, but also skillfully handles the complex dynamics of modern financial markets. By effectively predicting and responding to changes with great precision, our method strengthens financial risk management strategies paving the road toward a major advancement in predictive financial risk management.

The rest of this paper is organized as follows. Section~\ref{related} provides a review of existing literature on options hedging and the application of deep reinforcement learning in financial strategies, laying the groundwork for the innovations introduced in this work. Section~\ref{system} presents an overview of the system architecture and design objectives. Section~\ref{action} details the action space and reward design, essential for implementing effective hedging strategies. In Section~\ref{algorithm}, we introduce the Adaptive Nesterov Accelerated Distributional Deep Hedging (ANADDH) method, which combines advanced optimization techniques with distributional reinforcement learning to improve hedging performance. Section~\ref{exp} describes the experimental setup, leveraging diverse simulation environments and data configurations to capture a wide range of market dynamics, allowing for a rigorous performance evaluation of our approach. Finally, we summarize our contributions and discusses potential directions for future research in Section~\ref{conclusion}.

\section{Related Work}
\label{related}
The foundational theories of option pricing and hedging, such as the Black-Scholes and Merton models, have established essential frameworks for valuing financial derivatives and guiding risk management strategies. However, these models encounter significant challenges in adapting to the dynamic and rapidly evolving conditions of modern financial markets~\cite{black1976pricing}\cite{merton1973theory}. Advanced computational methods were developed to better address the uncertainties and volatilities in financial trading systems~\cite{hull2016options}\cite{zhao2023federated}.

Dynamic Vega hedging, crucial for managing volatility risks in financial derivatives, has evolved with the increasing application of machine learning techniques, particularly RL~\cite{cao2023gamma}. This transition highlights a significant shift from traditional methods to strategies capable of adapting to complex market dynamics~\cite{kling2011impact}.

Recent advancements in reinforcement learning have expanded its utility in financial markets, allowing for the development of sophisticated models that learn and adapt through iterative interactions with market data. The integration of deep learning with RL has further enhanced the capability of these models to process large volumes of data and extract actionable insights for effective risk management~\cite{jiang2017deep}\cite{ moody2001learning}.

A key innovation in this area is the application of distributional perspectives to RL, which models the entire distribution of possible outcomes rather than just the expected values~\cite{huang2022distributional}\cite{duan2021distributional}. This approach provides a comprehensive understanding of potential risks and returns, facilitating more robust financial decision-making~\cite{bellemare2017distributional}. Implicit Quantile Networks, which utilize quantile regression within distributional RL frameworks, exemplify this advancement by enabling the estimation of potential financial outcomes without predefined distributional assumptions~\cite{dabney2018implicit}\cite{bai2022monotonic}.

Furthermore, the field has seen significant enhancements in algorithm efficiency and effectiveness, particularly through the development of algorithms like Distributional Policy Optimization, which optimize policy decisions in complex, continuous control tasks common in financial applications~\cite{tessler2019distributional}. These innovations underscore the growing importance of adaptive and sophisticated computational techniques in the effective management of financial risks.

Despite recent advancements in applying reinforcement learning to hedging, the integration of cutting-edge RL methods with sophisticated optimization techniques, especially for Vega hedging, remains relatively unexplored. Existing work often lacks the adaptability required to handle the rapid fluctuations and tail risks characteristic of modern markets, where traditional RL methods may struggle to achieve stable, efficient performance. Key open challenges include the need for faster convergence, heightened sensitivity to volatility shifts, and improved accuracy in risk management across a range of possible outcomes. Addressing these gaps, this paper introduces an innovative approach that combines distributional RL with adaptive Nesterov acceleration, aiming to enhance stability, precision, and responsiveness in Vega hedging applications.

\begin{table}[t]
	\caption{Description of Notations in Training Procedure}
	\label{sample-table}
	\begin{center}
		\begin{tabular}{ll}
			\multicolumn{1}{c}{\bf NOTATION}  &\multicolumn{1}{c}{\bf DESCRIPTION}
			\\ \hline \\
			$\mathcal{S}$ & Set of states\\
			$\mathcal{A}$ & Set of actions\\
			$\kappa$      &  The transaction cost associated with the option used for hedging\\
			$p(s' | s, a)$ & Transition probability from state $s$ to state $s'$ under action $a$.\\
			$\mathcal{D}$ & Replay buffer containing experiences $(s, a, r, s')$\\
			$H_{\tau}(\delta)$& Huber loss function with parameter $\tau$\\
			$\theta$ & Parameters of the distributional RL model.\\
			$\mathcal{L}(\theta)$ & Loss function parameterized by $\theta$.\\
			$\tau$ & Index representing the quantile level.\\
			$\gamma$ & Discount factor for future rewards.\\
			$Z_{\tau}$& Target distribution for the $\tau$-th quantile\\
			$\delta_{\tau}$ & Temporal difference error for the $\tau$-th quantile \\
			$\pi(s)$ & Softmax policy for state $s$ parameterized by $\theta$.
			\\ \hline
		\end{tabular}
	\end{center}
\end{table}

\section{System Architecture and Design Objectives}
\label{system}
The Adaptive Nesterov Accelerated Distributional Deep Hedging (ANADDH) framework offers a robust, adaptive solution for managing Vega risk within a hedging portfolio of options by dynamically responding to volatility changes. Leveraging a distributional RL approach within a critic-actor network architecture, ANADDH captures the distribution of cumulative rewards to assess the overall performance of its hedging strategy effectively. The framework is built around two core components, i.e., the actor network, which selects optimal hedging actions, and the critic network, which evaluates these actions by modeling potential outcomes in terms of cumulative risk and cost.

ANADDH leverages both real-time and historical data inputs to continuously inform its decision-making. Key inputs include implied and realized volatility metrics to measure Vega sensitivity, essential Greeks, such as Delta and Vega, to assess price sensitivity to market conditions, and broader market indicators including price trajectories, trading volume, and transaction costs. This flow of data continuously updates the framework, enabling the actor network to make informed, data-driven hedging decisions, while the critic network evaluates a diverse spectrum of potential outcomes under varying market scenarios.

Within the ANADDH framework, the actor and critic networks function in a tightly integrated feedback loop. The actor network initiates hedging actions based on real-time market observations, while the critic network evaluates these actions by modeling the distribution of cumulative rewards. This iterative feedback system empowers the actor network to continuously refine its policy based on the critic's evaluations, resulting in a progressively optimized hedging strategy. Central to this process, adaptive Nesterov acceleration enhances the stability and efficiency of the training procedure by predicting future objective landscape information through momentum-driven foresight that leverages both current and historical gradient information. This forecasting enables precise, adaptive updates that anticipate shifts in the market landscape, aligning parameter adjustments with emerging financial conditions.

Our design prioritizes efficient convergence by discouraging unnecessary trading actions, stabilizing returns through controlled updates, and dynamically adjusting hedging strategies in response to shifting volatility. Together, these mechanisms allow ANADDH to address the primary challenges of traditional hedging methods, offering a resilient and adaptive approach to managing Vega risk effectively across diverse and dynamic market environments.
Detailed formulations and designs of the action and reward structures, along with the specifics of adaptive Nesterov acceleration within the distributional RL setup for dynamic Vega hedging, are discussed in the following Section~\ref{action} and Section~\ref{algorithm}.

\section{Action and Reward Strategy Design}
\label{action}
\subsection{Action Space Configuration}
The design of the action space is pivotal for the success of our dynamic Vega hedging strategy. At each time step $ t $, the action $ a_t^i $ specifies the hedging level as a proportion of the permissible maximum, described by the constraint
\begin{equation}
	\begin{aligned}
		a_t^i \in [0, a_{\text{max}}^i],
	\end{aligned}
\end{equation}
where $ a_{\text{max}}^i $ denotes the upper bound for hedging actions for option $i$. This limit is derived from the Vega sensitivity of the option, which measures the sensitivity of the option's price to a one-percentage-point change in the volatility of the underlying asset.

To establish $ a_{\text{max}}^i $, we first calculate the total Vega for the portfolio, summing the Vega values of all individual options. This total Vega is essential for gauging the portfolio's overall vulnerability to volatility shifts. We then determine the portfolio's risk tolerance, defined as the maximum acceptable change in portfolio value due to volatility movements. $ a_{\text{max}}^i $ is set by equating this risk tolerance to the total Vega, ensuring that each option's hedging action is appropriately scaled according to current market volatility conditions. This process ensures that our hedging actions are scaled in proportion to the portfolio's estimated volatility risk, based on the recent market dynamics reflected in the calculated total Vega. Although this approach uses the past volatility conditions as a benchmark, $a_{\text{max}}^i$ is continuously updated to adjust the hedging levels as new market data becomes available, allowing the strategy to adapt to changes in current and anticipated volatility.

\subsection{Hedging Efficiency Reward Design}
The reward function in our Vega hedging strategy framework is designed to promote portfolio stability and cost efficiency, shaping what we refer to as the hedging efficiency distribution within the distributional RL framework,
\begin{equation}
	\begin{aligned}
		r_t = r_t^{\text{PNL}} + r_t^{\text{cost}},
	\end{aligned}
\end{equation}
where $ r_t^{\text{PNL}} $ addresses portfolio stability by penalizing large fluctuations in value. This component is defined as the negative absolute difference in Profit and Loss (PNL) between consecutive time steps
\begin{equation}
	\begin{aligned}
		r_t^{\text{PNL}} = -|P_t - P_{t-1}|,
	\end{aligned}
\end{equation}
which encourages the development of hedging strategies that minimize significant market swings, thereby maintaining more stable portfolio values.

Additionally, the reward function includes a penalty for transaction costs associated with trading actions, designed to discourage unnecessary trading
\begin{equation}
	\begin{aligned}
		r_t^{\text{cost}} = -\sum_{i=1}^{N} c_i \cdot |a_t^i|,
	\end{aligned}
\end{equation}
where $ c_i $ represents the transaction cost per unit for trading option $ i $, and $ |a_t^i| $ quantifies the magnitude of the trading action for each option. This cost-related component of the reward function motivates the agent to optimize trading activities strategically, aiming to reduce unnecessary transaction expenses and enhance overall cost efficiency of the hedging strategy.
For simplicity in our upcoming formulations, we will use the notation $a$ to represent general hedging actions.

The reward design focuses on promoting efficient hedging by balancing portfolio stability and transaction cost reduction. While this approach does not explicitly target substantial profit maximization, it ensures that hedging actions are risk-conscious and cost-effective. It is important to recognize that the primary goal of our design is to control risk efficiently within the hedging portfolio, rather than to maximize profits. The profit maximization objective is intended for the investment portfolio, whereas our hedging strategy aims to mitigate risks and stabilize returns. Therefore, the reward mechanism prioritizes risk-adjusted performance and cost-effective risk control over pursuing higher profit levels. This aligns with the principles of efficient hedging, where the main focus is on minimizing downside risk and maintaining consistent portfolio behavior.

\section{Adaptive Nesterov Acceleration in Distributional RL for Dynamic Vega Hedging}
\label{algorithm}
In this section, we present the core components of  ANADDH, which incorporates adaptive Nesterov acceleration within a distributional reinforcement learning framework. The update direction in ANADDH is calculated using both current gradient information and momentum from previous updates. This momentum effectively looks ahead by predicting the likely movement of parameters based on past trends, enabling preemptive corrections and adjustments to the parameter path. ANADDH aimes to enhance the stability and precision of dynamic Vega hedging, allowing for adaptive adjustments that respond effectively to varying landscape complexities in the optimization process.

\subsection{Objective Function Design in ANADDH}
\subsubsection{Critic Network Objective}
The critic network serves as the analytical core of our ANADDH, tasked with assessing the potential value of actions taken by the actor network. This evaluation is achieved through a sophisticated modeling of the full distribution of possible accumulated rewards under the adopted policy, formalized as
\begin{equation}
	\begin{aligned}
		Z^{\pi}(s, a) \triangleq \sum_{t=0}^{\infty} \gamma^t r_{t+1} \mathbb{I}_{(s_t=s, a_t=a)},
	\end{aligned}
\end{equation}
where $\gamma$ is the discount factor, $r_{t+1}$ the reward received at time $ t+1 $, and $\mathbb{I}_{(s_t=s, a_t=a)}$ is an indicator function isolating accumulated rewards directly linked to the hedging action $a$ taken in state $s$. This rigorous approach enables the critic to capture a broad spectrum of potential outcomes for each state-action pair, significantly enhancing its capability to evaluate and manage risks efficiently.
The actor network's decisions, represented by the continuous action $a$, are evaluated by the critic. This evaluation is integral to refining the policy, directing the hedger towards optimal hedging strategies. 

In our ANADDH, the critic network evaluates each state-action pair by predicting the distribution of cumulative rewards across various quantile levels
\begin{equation}
	\begin{aligned}
		Z_{\tau}(s, a; \theta_{\text{critic}}) = f_{\tau}(s, a; \theta_{\text{critic}}),
	\end{aligned}
\end{equation}
where $\theta_{\text{critic}}$ represents the parameters of the critic network that shape the predicted quantile values. As the actor network refines its strategy, the range of actions evaluated by the critic also evolves, fostering a continuously adaptive estimation process that is crucial for dynamically adjusting the hedging strategy in response to market dynamics.
To capture the forward-looking nature of financial decision-making, the target distribution for each quantile is established as
\begin{equation}
	\begin{aligned}
		T_{\tau}(s, a) = r + \gamma Z_{\tau}(s', \pi(s'); \theta_{\text{critic}}),
	\end{aligned}
\end{equation}
where $s'$ denotes the subsequent state resulting from the current state $s$ and action $a$, and $ \pi(s') $ is the action determined by the actor's policy for that future state. This target distribution formulation directly correlates with the actor network's decisions, reinforcing the synergy between actor and critic in refining prediction accuracy.
The discrepancy between the predicted and target quantiles, $ \delta_{\tau}(s, a) $, is defined as
\begin{equation}
	\begin{aligned}
		\delta_{\tau}(s, a) = Z_{\tau}(s, a; \theta_{\text{critic}}) - T_{\tau}(s, a),
	\end{aligned}
\end{equation}
which quantifies the error in the critic's predictions relative to the expected outcomes based on the actor's policy and resulting state transitions.
This discrepancy is essential for calculating the Quantile Huber Loss, guiding updates to the critic's parameters to reduce prediction inaccuracies and enhance the overall reliability and effectiveness of the hedging strategy.

The precision of the critic network's predictions is fine-tuned using the Quantile Huber Loss function, designed to balance sensitivity to minor prediction errors with robustness against outliers
\begin{equation}
	\begin{aligned}
		\mathcal{L}(\theta_{\text{critic}}) = \sum_{\tau \in \mathcal{T}} \mathbb{E}_{(s,a) \sim \mathcal{D}} \left[\rho_{\tau} \cdot H_{\tau}(\delta_{\tau}(s,a))\right],
	\end{aligned}
\end{equation}
where $\rho_{\tau}$ adjusts for the importance of each quantile level, and $H_{\tau}$ defines the Quantile Huber Loss applied to the discrepancies between predicted and target quantiles
\begin{equation}
	\begin{aligned}
		H_{\tau}(\delta_{\tau}(s, a)) = 
		\begin{cases} 
			\frac{1}{2}(\delta_{\tau}(s, a))^2 & \text{if } |\delta_{\tau}(s, a)| \leq \kappa, \\
			\kappa(|\delta_{\tau}(s, a)| - \frac{1}{2}\kappa) & \text{if } |\delta_{\tau}(s, a)| > \kappa,
		\end{cases}
	\end{aligned}
\end{equation}
with $\kappa$ acting as the threshold for transitioning from quadratic to linear loss. This loss function enables the critic to effectively minimize prediction errors across all quantiles while maintaining resilience against disruptive market anomalies.

\subsubsection{Actor Network Objective}
The actor network's objective function is defined within the policy function, parameterized by $ \theta_{\text{actor}} $ as $ \pi(s; \theta_{\text{actor}}) $. This function outputs a continuous action $ a $ based on the state $ s $, using a neural network model
\begin{equation}
	\begin{aligned}
		a = f(s; \theta_{\text{actor}})
	\end{aligned}
\end{equation}
where $f(s; \theta_{\text{actor}})$ represents the neural network's output, detailing the specific action to be taken in state $s$, such as adjusting the proportion of maximum hedging. This deterministic mapping enables precise and real-time hedging decisions based on the actor's assessment of market conditions.

The actor network's primary goal is to maximize the expected return under the policy parameterized by $\theta_{\text{actor}}$
\begin{equation}
	\begin{aligned}
		J(\theta_{\text{actor}}) = \mathbb{E}_{s \sim p_{\pi_{\theta_{\text{actor}}}}} \left[ \sum_{t=0}^{\infty} \gamma^t r_t \right],
	\end{aligned}
\end{equation}
where $\gamma$ is the discount factor, and $r_t$ is the reward at time $t$. The optimization of this expected return is typically pursued through gradient ascent on $J(\theta_{\text{actor}})$, which systematically adjusts $\theta_{\text{actor}}$ to enhance policy performance.

The actor network's decisions are constrained to ensure actions remain within specified ranges, such as maintaining the Vega ratio after hedging within predefined bounds from $[0, 1]$. This constraint is implemented through an activation function in the neural network's final layer, directing the actor's decisions towards optimal hedging strategies and away from speculative trading. These constraints not only improve sample efficiency by directing the agent toward more relevant actions but also enhance convergence by limiting the action space to a well-defined range, simplifying the learning environment.

Each action selection by the actor prompts an evaluation by the critic network, which assesses the potential value of these actions given the current state
\begin{equation}
	\begin{aligned}
		Z(s, a; \theta_{\text{critic}}) = f_{\text{critic}}(s, a; \theta_{\text{critic}}).
	\end{aligned}
\end{equation}
This value assessment forms the basis for computing the policy gradient, which is used to adjust the actor's policy parameters. The feedback from these evaluations dynamically refines the actor's strategy, guiding adjustments to actions projected to yield higher returns and ensuring the hedging strategy remains aligned with evolving market conditions.

\subsection{Gradient Information in ANADDH}
\subsubsection{Gradient Calculation of Critic Network}
The implementation of ANA within the critic network's update mechanism leverages gradient information derived from both the actor and critic networks, optimizing parameter updates. This integration forms a dynamic feedback loop that is crucial for continuous learning and the enhancement of policies.

The actor network selects actions based on its policy $\pi(a \mid s)$, defining the future state-action pairs that the critic network will evaluate. These evaluations are essential for refining the value estimates, which are crucial for the actor's subsequent policy updates. The ongoing interaction ensures that both networks are synchronized in their learning objectives, enhancing the effectiveness of the hedging strategy.

The overall objective function for the critic network, denoted by $\mathcal{L}(\theta_{\text{critic}})$, incorporates the expected gradients of the Quantile Huber Loss
\begin{equation}
	\nabla_{\theta_{\text{critic}}} \mathcal{L}(\theta_{\text{critic}}) = \sum_{\tau \in \mathcal{T}} \mathbb{E}_{(s,a) \sim \mathcal{D}} \left[\rho_{\tau} \cdot \nabla_{\theta_{\text{critic}}} H_{\tau}(\delta_{\tau}(s,a))\right],
\end{equation}
where the gradient of the Quantile Huber Loss depends on the discrepancy between predicted and target quantiles, and it adjusts the critic's parameters to reduce prediction errors effectively.

For smaller prediction errors as $|\delta_{\tau}(s, a)| \leq \kappa$, where the loss function is quadratic, the gradient is calculated as
\begin{equation}
	\begin{aligned}
		\nabla_{\theta_{\text{critic}}} H_{\tau}(\delta_{\tau}(s, a)) = \delta_{\tau}(s, a) \cdot \nabla_{\theta_{\text{critic}}} \delta_{\tau}(s, a),
	\end{aligned}
\end{equation}
reflecting a typical gradient descent update where the adjustment magnitude is proportional to the error.
For larger errors as $|\delta_{\tau}(s, a)| > \kappa$, where outliers might distort learning, the loss function becomes linear to mitigate their influence, and the gradient is given by
\begin{equation}
	\begin{aligned}
		\nabla_{\theta_{\text{critic}}} H_{\tau}(\delta_{\tau}(s, a)) = \kappa \cdot \text{sign}(\delta_{\tau}(s, a)) \cdot \nabla_{\theta_{\text{critic}}} \delta_{\tau}(s, a),
	\end{aligned}
\end{equation}
ensuring that the gradient's magnitude remains constant $\kappa$, which provides stability and robustness in the presence of large errors.

The actor network's decisions directly influence the target quantile values $T_{\tau}(s, a)$, which are adjusted based on the actor's future policy decisions
\begin{equation}
	\nabla_{\theta_{\text{critic}}} T_{\tau}(s, a) = \gamma \nabla_{\theta_{\text{critic}}} Z_{\tau}(s', \pi(s'); \theta_{\text{critic}}),
\end{equation}
where $s'$ is the subsequent state determined by the actor's policy $\pi(s')$. These decisions shape the future state-action pairs evaluated by the critic, impacting the critic's parameter updates through the gradients
\begin{equation}
	\nabla_{\theta_{\text{critic}}} \delta_{\tau}(s, a) = \nabla_{\theta_{\text{critic}}} Z_{\tau}(s, a; \theta_{\text{critic}}) - \nabla_{\theta_{\text{critic}}} T_{\tau}(s, a).
\end{equation}
This enhanced feedback mechanism ensures that the critic's learning is closely aligned with the actual policy performance, enabling precise adjustments to optimize the overall hedging strategy in response to evolving market conditions.

\subsubsection{Gradient Calculation of Actor Network}
Policy gradients are fundamental for optimizing hedging strategies within our framework. They quantify the rate of change of the expected return with respect to the policy network parameters, $ \theta_{\text{actor}} $, reflecting the cumulative financial performance of the hedging actions over time.

The primary objective of the policy network is to identify optimal hedging actions by learning a policy that maximizes expected rewards. The policy gradients facilitate this by directing parameter adjustments in the actor network to enhance the expected return. The gradient of the objective function with respect to the policy parameters is formulated as
\begin{equation}
	\nabla_{\theta_{\text{actor}}} J(\theta_{\text{actor}}) = \sum_{t=0}^{\infty} \nabla_{\theta_{\text{actor}}} \mathbb{E}_{\pi_{\theta_{\text{actor}}}} \left[ \gamma^t r_t \right].
\end{equation}
The computation of this gradient is derived using the policy gradient theorem as
\begin{equation}
	\begin{aligned}
		\nabla_{\theta_{\text{actor}}} \mathbb{E}_{\pi_{\theta_{\text{actor}}}} \left[ \gamma^t r_t \right] = \mathbb{E}_{\pi_{\theta_{\text{actor}}}} \left[ \nabla_{\theta_{\text{actor}}} \log(\pi(a_t|s_t; \theta_{\text{actor}})) \cdot \gamma^t r_t \right],
	\end{aligned}
\end{equation}
where $ \log(\pi(a_t|s_t; \theta_{\text{actor}})) $ denotes the log-probability of selecting action $ a_t $ given state $ s_t $ under the policy parameterized by $ \theta_{\text{actor}}$.
The gradient of the log-probability is expressed as
\begin{equation}
	\nabla_{\theta_{\text{actor}}} \log(\pi(a_t|s_t; \theta_{\text{actor}})) = \frac{\nabla_{\theta_{\text{actor}}} \pi(a_t|s_t; \theta_{\text{actor}})}{\pi(a_t|s_t; \theta_{\text{actor}})},
\end{equation}
which effectively decomposes the gradient into the reciprocal of the action's probability and the gradient of the policy output with respect to the parameters.
By summing over all time steps, the overall gradient for updating the policy is given by
\begin{equation}
	\begin{aligned}
		\nabla_{\theta_{\text{actor}}} J(\theta_{\text{actor}}) = \sum_{t=0}^{\infty} \mathbb{E}_{\pi_{\theta_{\text{actor}}}} \left[ \nabla_{\theta_{\text{actor}}} \log(\pi(a_t|s_t; \theta_{\text{actor}})) \cdot \gamma^t r_t \right],
	\end{aligned}
\end{equation}
which enables a comprehensive update to the actor network's policy by summing over all time steps, incorporating both immediate and future rewards. By leveraging the energy of the gradient, this approach not only refines the actor's policy continuously but also provides insight into the curvature of the loss landscape. This curvature information is critical for estimating changes in the objective function and paves the way for predictive updates in ANADDH. By aligning updates with the anticipated trajectory of market dynamics, this method enhances the actor network's ability to manage financial risks and improve the robustness of portfolio performance.

\subsection{Predictive Updating in ANADDH}
The proposed ANADDH framework significantly improves training dynamics within the actor-critic structure by integrating an advanced optimization mechanism that leverages gradient energy to estimate the curvature of the objective function. This curvature estimation, combined with accumulated momentum from previous updates, enables adaptive and stable convergence for both the critic and actor networks.

ANADDH initiates with a predictive step that leverages momentum to forecast future gradients, informed by the estimated curvature of the objective function. This forecasting enables proactive adjustments to parameters, allowing the model to anticipate shifts in the dynamic financial environment. Following this, a corrective update fine-tunes these adjustments, ensuring they remain precise and responsive to the latest market conditions.

Central to ANADDH is the establishment of an auxiliary point for each network, serving as a forward-looking estimate within the parameter update trajectory and incorporating both past momentum and curvature information. This enhanced foresight significantly boosts the predictive accuracy and efficiency of the optimization process, enabling the networks to make targeted adjustments that align with anticipated future states and sustain readiness for evolving market dynamics.

Both the critic and actor networks use Nesterov's momentum to independently adapt their updates based on past gradients, yet in a way that aligns with the framework's overarching objective. Each network follows the momentum adjustment rule
\begin{equation}
	t_{r+1} = \frac{1 + \sqrt{1 + 4t_r^2}}{2},
\end{equation}
where $t_1 = 1$. This formula dynamically tunes momentum parameters, refining the learning trajectory for both networks by integrating insights from both current and historical gradient information. Although the networks are updated asynchronously and have distinct objectives, the shared adaptation process allows them to progress in harmony, promoting stable and efficient optimization within the ANADDH framework.

The predictive updating in ANADDH initializes by defining an auxiliary point, which serves as a reference in the updating trajectory of the critic network's parameters
\begin{equation}
	\bm{y_r^{\text{critic}}} = \bm{\theta_{\text{critic}}^r} + \frac{t_r - 1}{t_{r+1}} (\bm{\theta_{\text{critic}}^r} - \bm{\theta_{\text{critic}}^{r-1}}),
\end{equation}
which not only guides the direction toward minimizing the objective function $\mathcal{L}(\bm{\theta_{\text{critic}}})$, but also integrates momentum information by utilizing gradient moments calculated at the auxiliary point.

The first moment, $\bm{m_r^{\text{critic}}}$, captures the directional gradient information, accumulating momentum along the primary optimization path to support smooth convergence, formulated as
\begin{equation}
	\bm{m_r^{\text{critic}}} = \beta_1 \bm{m_{r-1}^{\text{critic}}} + (1 - \beta_1) \bm{g_s(y_r^{\text{critic}})},
\end{equation}
which is regulated by the decay rate $ \beta_1 $ to control how quickly the first moment adapts to changes in the gradient's direction. The second moment, $\bm{v_r^{\text{critic}}}$, approximates the gradient's energy, serving as an adaptive estimate of the curvature information of the objective function as
\begin{equation}
	\bm{v_r^{\text{critic}}} = \beta_2 \bm{v_{r-1}^{\text{critic}}} + (1 - \beta_2) \bm{g_s(y_r^{\text{critic}})}^2,
\end{equation}
where decay rate $\beta_2$ tunes the responsiveness of this second moment, allowing it to reflect shifts in gradient energy and curvature effectively. 

Following the calculation of the adaptive direction using $ \bm{y_r^{\text{critic}}} $, an accelerated update is applied
\begin{equation}
	\bm{\theta_{\text{critic}}^r} = \bm{y_r^{\text{critic}}} - \frac{\sqrt{1 - \beta_2^{r}}}{1 - \beta_1^{r}} \frac{\bm{m_r^{\text{critic}}}}{\sqrt{\bm{v_r^{\text{critic}}} + \epsilon}},
\end{equation}
where $ \epsilon $ is a small constant ensuring numerical stability. This formulation integrates both the directional path and adaptive curvature information from the gradient, allowing the update to be finely tuned in response to the landscape of the objective function. By adjusting the update magnitude based on both the gradient direction and estimated curvature, ANADDH achieves a highly responsive and stable adjustment mechanism that aligns with changing financial market dynamics. This Nesterov-based adaptive update represents a core innovation in ANADDH, facilitating accelerated learning that anticipates and adjusts to shifts in market conditions, enhancing both precision and robustness in parameter optimization.

In the actor network, ANADDH employs adaptive Nesterov updating by integrating momentum and adaptive learning rates to refine parameter updates. The first moment, $ \bm{m_r^{\text{actor}}} $, accumulates directional information, ensuring smooth policy updates, while the second moment, $ \bm{v_r^{\text{actor}}} $, estimates the gradient's energy to adaptively capture curvature information. These moments are updated as
\begin{equation}
	\bm{m_r^{\text{actor}}} = \beta_1 \bm{m_{r-1}^{\text{actor}}} + (1 - \beta_1) \nabla_{\theta_{\text{actor}}} J(\theta_{\text{actor}}),
\end{equation}
\begin{equation}
	\bm{v_r^{\text{actor}}} = \beta_2 \bm{v_{r-1}^{\text{actor}}} + (1 - \beta_2) (\nabla_{\theta_{\text{actor}}} J(\theta_{\text{actor}}))^2.
\end{equation}
Using these, the adaptive learning rate $ \alpha_r^{\text{actor}} $ is calculated as
\begin{equation}
	\alpha_r^{\text{actor}} = \frac{\sqrt{1 - \beta_2^r}}{1 - \beta_1^r} \cdot \frac{\bm{m_r^{\text{actor}}}}{\sqrt{\bm{v_r^{\text{actor}}} + \epsilon}}.
\end{equation}
An auxiliary point $ \bm{y_r^{\text{actor}}} $ is then computed to anticipate the trajectory of future updates
\begin{equation}
	\bm{y_r^{\text{actor}}} = \bm{\theta_{\text{actor}}^r} + \frac{t_r - 1}{t_{r+1}} (\bm{\theta_{\text{actor}}^r} - \bm{\theta_{\text{actor}}^{r-1}}).
\end{equation}
The final parameter update for the actor network is applied as
\begin{equation}
	\bm{\theta_{\text{actor}}^r} = \bm{y_r^{\text{actor}}} - \alpha_r^{\text{actor}} \cdot \nabla_{\theta_{\text{actor}}} J(\theta_{\text{actor}}).
\end{equation}
The proposed adaptive approach enables the actor network to leverage both current and historical gradient information, aligning updates with the anticipated trajectory of policy changes, thereby enhancing stability and responsiveness within the complex financial landscape addressed by ANADDH.

\subsection{Efficiency Analysis of Predictive Updating in ANADDH}
The proposed ANADDH framework enhances training dynamics within the actor-critic structure by leveraging both real-time gradient information and accumulated momentum from prior updates. This integration allows ANADDH to not only anticipate the trajectory of parameter updates but also adjust them proactively, ensuring that each step aligns closely with the optimal path. By predicting and correcting for potential deviations, i.e., whether oversteps or insufficient adjustments, our approach enhances both the stability and precision of the learning process, enabling the framework to respond dynamically to the complex landscape of financial market changes.

\subsubsection{Quadratic Approximation}
To analyze the optimality of the predictive step in ANADDH, we employ a linear interpolation between the auxiliary and updated parameter points which is defined as
\begin{equation}
	\bm{z_{\text{critic}}} = \bm{y_r^{\text{critic}}} + \alpha (\bm{\theta_{\text{critic}}} - \bm{y_r^{\text{critic}}}),
\end{equation}
where $\alpha$ is a scalar parameter ranging from 0 to 1, controlling the gradual shift from the auxiliary point $\bm{y_r^{\text{critic}}}$ to the updated parameter point $\bm{\theta_{\text{critic}}}$. This controlled transition facilitates a systematic exploration of the parameter space along the optimization trajectory, helping to refine the accuracy of each update.

The effectiveness of ANADDH's predictive updating is evaluated by examining how variations in the parameter $ \alpha $ impact the objective function $ \mathcal{L} $. This sensitivity is captured by the first-order derivative of $ \mathcal{L}(\bm{z_{\text{critic}}}) $ with respect to $ \alpha $, which serves as a guide for the update trajectory
\begin{equation}
	\frac{d\mathcal{L}(\bm{z_{\text{critic}}})}{d\alpha} = \nabla \mathcal{L}(\bm{z_{\text{critic}}})^T (\bm{\theta_{\text{critic}}} - \bm{y_r^{\text{critic}}}),
\end{equation}
offering insights into the behavior of $ \mathcal{L} $ along the interpolation path defined by $ \alpha $. 
To evaluate the cumulative impact of the parameter updates on $ \mathcal{L} $, we integrate this derivative from 0 to 1 as
\begin{equation}
	\int_{0}^{1} \frac{d\mathcal{L}(\bm{z_{\text{critic}}})}{d\alpha} d\alpha = \int_{0}^{1} \nabla \mathcal{L}(\bm{z_{\text{critic}}})^T (\bm{\theta_{\text{critic}}} - \bm{y_r^{\text{critic}}}) d\alpha,
\end{equation}
which leads to the integral result as
\begin{equation}
	\mathcal{L}(\bm{\theta_{\text{critic}}}) - \mathcal{L}(\bm{y_r^{\text{critic}}}) = \int_{0}^{1} \nabla \mathcal{L}(\bm{z_{\text{critic}}})^T (\bm{\theta_{\text{critic}}} - \bm{y_r^{\text{critic}}}) d\alpha,
\end{equation}
capturing the total influence of parameter updates on the objective function.

The integration and application of the Cauchy-Schwarz inequality are essential in establishing bounds on changes in the objective function during the critic network's update process within ANADDH. By applying the Cauchy-Schwarz inequality, we can derive an upper bound for the following expression
\begin{equation}
	\begin{aligned}
		\left| \int_{0}^{1} (\nabla \mathcal{L}(\bm{z_{\text{critic}}}) - \nabla \mathcal{L}(\bm{y_r^{\text{critic}}}))^T (\bm{\theta_{\text{critic}}} - \bm{y_r^{\text{critic}}}) d\alpha \right| \leq 
		\int_{0}^{1} \left\| \nabla \mathcal{L}(\bm{z_{\text{critic}}}) - \nabla \mathcal{L}(\bm{y_r^{\text{critic}}}) \right\|_2 \left\| \bm{(\theta_{\text{critic}} - y_r^{\text{critic}})} \right\|_2 d\alpha.
	\end{aligned}
\end{equation}
By defining the maximum eigenvalue of the Hessian of the objective function at $\bm{y_r^{\text{critic}}}$ as $L$, we can further simplify
\begin{equation}
	\left\| \nabla \mathcal{L}(\bm{z_{\text{critic}}}) - \nabla \mathcal{L}(\bm{y_r^{\text{critic}}}) \right\|_2 \leq \alpha L \left\| \bm{\theta_{\text{critic}} - y_r^{\text{critic}}} \right\|_2,
\end{equation}
which results in the integral being bounded by
\begin{equation}
	\begin{aligned}
		\int_{0}^{1} \left\| \nabla \mathcal{L}(\bm{z_{\text{critic}}}) - \nabla \mathcal{L}(\bm{y_r^{\text{critic}}}) \right\|_2 \left\| \bm{(\theta_{\text{critic}} - y_r^{\text{critic}})} \right\|_2 d\alpha \leq \frac{L}{2} \left\| \bm{(\theta_{\text{critic}} - y_r^{\text{critic}})} \right\|_2^2,
	\end{aligned}
	\label{eq-40}
\end{equation}
allowing us to control the impact of parameter updates on the objective function, and ensuring stability in the critic network's optimization process within ANADDH.

\subsubsection{Bounding Gradient Changes Using the Hessian Matrix}
In the critic network's adaptive update process, the Hessian matrix of the objective function, $ \mathcal{L}(\theta_{\text{critic}}) $, is used to quantify and bound the rate at which gradients change. The gradient at any point $ \bm{z_{\text{critic}}} $ along the update path can be approximated through a Taylor series expansion as 
\begin{equation}
	\bm{\nabla \mathcal{L}(\bm{z_{\text{critic}}})} = \bm{\nabla \mathcal{L}(\bm{y_r^{\text{critic}}})} + \bm{\nabla^2 \mathcal{L}(\bm{y_r^{\text{critic}}})} (\bm{z_{\text{critic}}} - \bm{y_r^{\text{critic}}}),
\end{equation}
where the Hessian, $ \bm{\nabla^2 \mathcal{L}(\bm{y_r^{\text{critic}}})} $, captures the curvature of the objective function, indicating how sensitive the gradient is to changes in the parameter vector $ \bm{z_{\text{critic}}} $. The use of the Hessian in the analysis helps ensure that the adaptive updates account for the underlying structure of the optimization landscape.

To effectively bound the change in the gradients, we perform an eigen-decomposition of the Hessian matrix
\begin{equation}
	\bm{\nabla^2 \mathcal{L}(\bm{y_r^{\text{critic}}})} = \bm{U \Lambda U^T},
\end{equation}
where $ \bm{U} $ is an orthogonal matrix consisting of eigenvectors, and $ \bm{\Lambda} $ is a diagonal matrix containing the eigenvalues. Assuming $ L $ as the upper bound of these eigenvalues, we establish
\begin{equation}
	\begin{aligned}
		\bm{\|\nabla^2 \mathcal{L}(\bm{y_r^{\text{critic}}}) (\bm{z_{\text{critic}}} - \bm{y_r^{\text{critic}}})\|^2} = (\bm{z_{\text{critic}}} - \bm{y}_r^{\text{critic}})^T U \Lambda^2 U^T (\bm{z_{\text{critic}}} - \bm{y_r^{\text{critic}}}).
	\end{aligned}
\end{equation}
By defining $ \bm{v} = \bm{U^T (\bm{z_{\text{critic}}} - \bm{y_r^{\text{critic}}})} $, we then have	
\begin{equation}
	\bm{v^T \Lambda^2 v} = \sum_{i = 1}^N \lambda_i^2 v_i^2 \leq \lambda_{\text{max}}^2 \bm{v^T v},
\end{equation}
where $ \lambda_{\text{max}} $ is the largest eigenvalue in $ \bm{\Lambda} $. Given that
\begin{equation}
	\begin{aligned}
		\bm{v^T v} &= \bm{\|U^T (\bm{z_{\text{critic}}} - \bm{y_r^{\text{critic}}})\|^2} = \bm{(\bm{z_{\text{critic}}} - \bm{y_r^{\text{critic}}})^T U U^T (\bm{z_{\text{critic}}} - \bm{y_r^{\text{critic}}})} = \bm{\|\bm{z_{\text{critic}}} - \bm{y_r^{\text{critic}}}\|^2},
	\end{aligned}
\end{equation}
the inequality simplifies to
\begin{equation}
	\bm{\|\nabla^2 \mathcal{L}(\bm{y_r^{\text{critic}}}) (\bm{z_{\text{critic}}} - \bm{y_r^{\text{critic}}})\|} \leq \lambda_{\text{max}} \bm{\|\bm{z_{\text{critic}}} - \bm{y_r^{\text{critic}}}\|}.
\end{equation}
Finally, we deduce that the change in the gradient can be bounded by
\begin{equation}
	\begin{aligned}
		\bm{\left \|\nabla \mathcal{L}(\bm{z_{\text{critic}}}) - \nabla \mathcal{L}(\bm{y_r^{\text{critic}}}) \right \|_2} \leq L \bm{\left \|\bm{z_{\text{critic}}} - \bm{y_r^{\text{critic}}}\right \|_2} = \alpha L \bm{\left \|\theta_{\text{critic}} - \bm{y_r^{\text{critic}}}\right \|_2}.
	\end{aligned}
\end{equation}
The above analytical approach to bounding gradient changes ensures that updates to the critic's parameters are both stable and predictable, optimizing the training dynamics for improved learning outcomes.

\subsubsection{Deriving a Quadratic Bound for the Objective Function}
By integrating the gradient change bounds with previously established constraints, we derive a compact expression for the integral that represents the variation in the objective function along the update path, as shown in (\ref{eq-40}), which leads to a quadratic approximation of the objective function, denoted by $ \psi(\bm{\theta_{\text{critic}}}) $ as
\begin{equation}
	\begin{aligned}
		\psi(\bm{\theta_{\text{critic}}}) = \mathcal{L}(\bm{y_r^{\text{critic}}}) + \bm{\nabla \mathcal{L}(\bm{y_r^{\text{critic}}})}^T (\bm{\theta_{\text{critic}} - y_r^{\text{critic}}}) + \frac{L}{2} \|\bm{\theta_{\text{critic}} - y_r^{\text{critic}}}\|_2^2.
	\end{aligned}
\end{equation}

To establish the conditions for optimal parameter updates, we set the gradient of $ \psi(\bm{\theta_{\text{critic}}}) $ to zero
\begin{equation}
	\bm{\nabla \psi(\bm{\theta_{\text{critic}}})} = \bm{\nabla \mathcal{L}(\bm{y_r^{\text{critic}}})} + L (\bm{\theta_{\text{critic}} - y_r^{\text{critic}}}) = \bm{0},
\end{equation}
leading to the identification of the optimal update direction
\begin{equation}
	\bm{\theta_{\text{critic}}^*} = \bm{y_r^{\text{critic}}} - \frac{1}{L} \bm{\nabla \mathcal{L}(\bm{y_r^{\text{critic}}})}.
\end{equation}
In ANADDH, we replace the current gradient $ \bm{\nabla \mathcal{L}(\bm{y_r^{\text{critic}}})} $ with adaptively updated gradient information, improving the robustness of our approach. By applying a quadratic bound derived from the adaptive momentum updates, the optimization landscape in ANADDH becomes simpler and more manageable, allowing for an efficient predictive updating process. While this bound does not guarantee convergence to a globally optimal solution, it serves as a tool for evaluating the efficiency of ANADDH's predictive updates. The quadratic approximation enables ANADDH to make well-informed adjustments by effectively estimating the curvature of the objective function's landscape, ensuring that parameter movements are both accurate and responsive to the shape of the optimization path.
This analysis also applies to the actor network's update procedures, enhancing the overall adaptive capabilities of the framework within the actor-critic model.

ANADDH combines adaptive Nesterov acceleration with distributional reinforcement learning, optimizing both the speed and stability of parameter updates for the critic and actor networks. By leveraging gradient moments, the framework estimates directional and curvature information, allowing it to dynamically adjust its learning trajectory according to the objective landscape. Through a rigorous quadratic approximation and eigenvalue-based bounding, ANADDH ensures stable convergence while minimizing prediction errors and transaction costs. This analysis confirms that ANADDH provides a robust and efficient approach for managing financial risks in volatile market conditions.

\section{Experiments}
\label{exp}
In this section, we provide the experimental framework designed to evaluate the performance of various hedging strategies using plain vanilla European call options. The algorithms compared include Delta Hedging~\cite{alexander2023delta}, Delta-Vega Hedging~\cite{herrmann2017model}, Gamma-Vega Deep Hedging (GVDH)~\cite{cao2023gamma}, and our proposed ANADDH. Our experiments employ a diverse set of market models where the price dynamics of the underlying assets follow geometric Brownian motion without jumps. This selection covers a wide range of realistic market behaviors, allowing for an extensive analysis of the strategies under various scenarios. Changes in the hedged portfolio's value between decisions are modeled as approximately normally distributed, reflecting a spectrum of real-world market fluctuations. This comprehensive approach ensures that our simulations provide a detailed and representative insight into financial market dynamics, offering valuable findings that extend beyond the scope of specific empirical datasets.

\subsection{Experimental Setup}
In our experimental setup, we employ simulated data to rigorously evaluate the performance of our hedging strategy under a controlled and diverse range of market conditions as shown in~\cite{cao2023gamma}. Simulated data offers several key advantages, particularly in enabling precise customization of parameters such as volatility, price movements, and transaction costs, allowing for systematic testing across various scenarios that may not be adequately represented in historical datasets. By simulating a broad spectrum of market dynamics, we mitigate biases inherent in real-world data, providing a more comprehensive evaluation of the model's robustness and adaptability. This approach allows us to examine the hedging strategy's performance in extreme or rare market conditions, ensuring that the outcomes are generalizable and resilient across both typical and challenging financial landscapes.

\subsubsection{Trading Environment and Orders}
In our setup, each option in the portfolio is an At-The-Money (ATM) European call, providing a standardized basis for comparison across experiments. We simulate client orders arriving according to a Poisson process with an intensity of one per day, each involving an option on 100 units of the underlying asset. These orders have equal probabilities of being long or short, with each option valued at approximately \$60 based on the Black-Scholes formula.

\subsubsection{Market Modeling}
The market is modeled without jumps, assuming the underlying asset's price changes follow a normal distribution between hedging actions. The experimental design incorporates a stochastic volatility model with parameters $\upsilon$ for volatility variation and $\sigma$ for initial volatility, using the SABR model parameters $\beta = 0.5$ and $\rho = 0.2$. Both the annual dividend yield $q$ and the risk-free interest rate $r$ are set to zero for simplification. Proportional transaction costs relative to the option price are assumed to capture their significant impact on hedging strategy performance.

We employ the SABR model, as detailed in Appendix~\ref{A2}, to simulate asset prices and instantaneous volatility, integrating stochastic elements into both. This modeling generates a range of potential market scenarios, with asset prices updated based on expected returns, current volatility levels, and random shocks to mimic realistic market fluctuations. Volatility levels are dynamically adjusted considering prior levels and stochastic variations, ensuring that the simulations closely align with market expectations and offer a realistic framework for analyzing trading environments.

\subsubsection{Objective Functions for Performance Evaluation}
To evaluate the efficiency of the ANADDH hedging strategy, we employ three objective functions that capture distinct magnitudes of hedging performance, focusing on minimizing risk exposure and achieving portfolio stability. These metrics assess how effectively the hedging strategy manages fluctuations in PNL and mitigates extreme losses.

Firstly, the Mean-STD function measures the magnitude of variability in the portfolio's PNL by balancing the mean and standard deviation. By minimizing this function, the hedging strategy aims to stabilize PNL outcomes, reducing unnecessary volatility and aligning with the reward design's goal of discouraging large PNL fluctuations. Secondly, Value at Risk (VaR-95\%) captures the magnitude of potential losses within the worst 5\% of PNL scenarios. This function provides an upper bound on adverse PNL magnitudes, ensuring that most losses remain below this threshold. Minimizing VaR-95\% indicates that the hedging strategy is effectively managing risk under severe conditions, contributing to portfolio robustness.
Finally, Conditional Value at Risk (CVaR-95\%) measures the average magnitude of PNL within the worst 5\% of scenarios, offering insight into the scale of extreme losses beyond the VaR-95\% threshold. Lowering the CVaR-95\% magnitude reflects a hedging strategy that effectively mitigates severe risks, enhancing stability even under volatile market conditions.

\subsection{Performance Analysis Across Different Option Maturities}
\begin{figure}
	\centering
	\begin{tabular}{cccc} 
		\includegraphics[width = 0.45\linewidth]{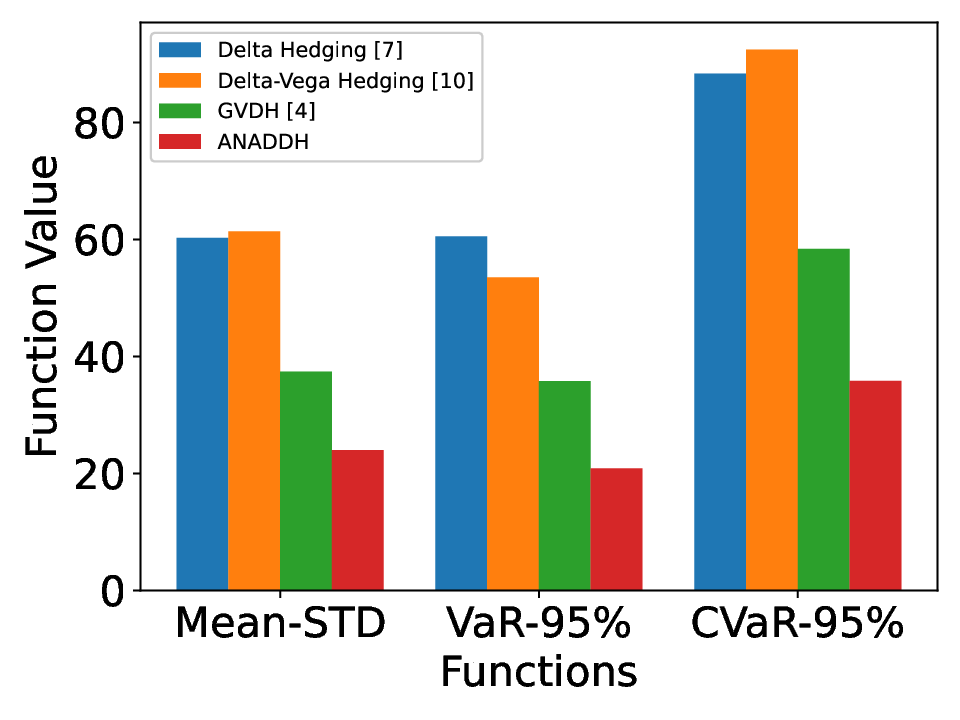} & 
		\includegraphics[width = 0.45\linewidth]{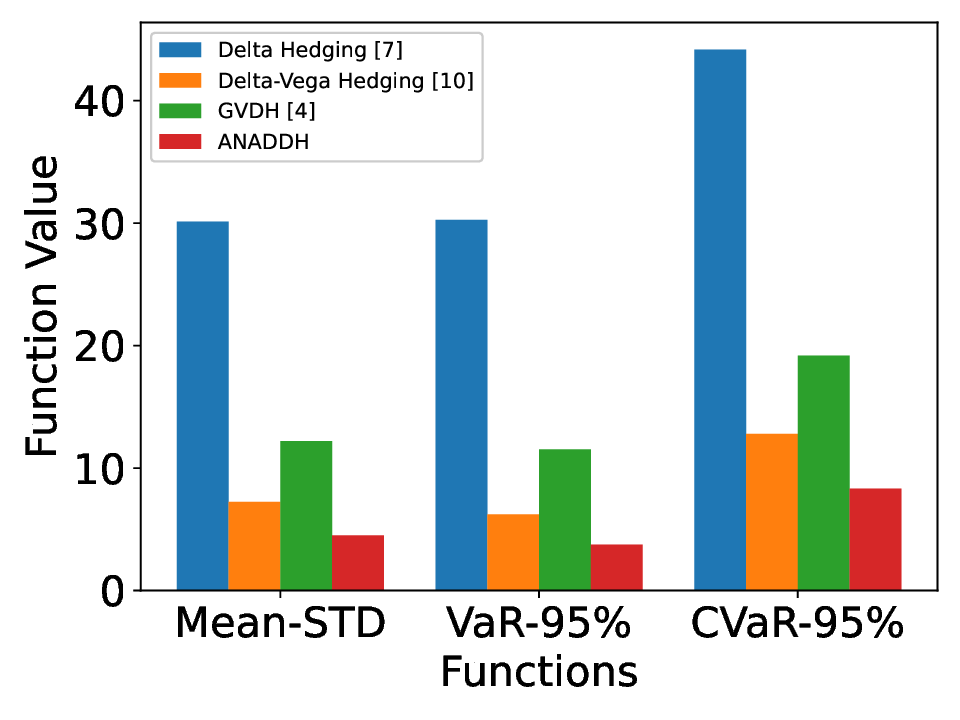} \\
		{\scriptsize (a) $30$ Days} &
		{\scriptsize (b) $120$ Days}  
	\end{tabular}
	\captionsetup{font={scriptsize}}
	\caption{Performance comparison with different hedging option maturity in days.}
	\label{fig-hedttm}
\end{figure}
In Fig.~\ref{fig-hedttm}, we compare the performance of four hedging strategies, i.e., Delta Hedging, Delta-Vega Hedging, GVDH, and our proposed ANADDH framework, across option maturities of 30 days and 120 days. For options with a 30-day maturity, shown in Fig.~\ref{fig-hedttm}(a), Delta Hedging and Delta-Vega Hedging display similar performance levels, which is expected given the high delta sensitivity of shorter maturity options. Delta Hedging proves effective with the 30-day maturity as it addresses the predominant delta-driven adjustments needed within short time frames. Although Delta-Vega Hedging adds a layer of volatility management, its benefit is marginal due to the limited time horizon. GVDH and ANADDH, which incorporate higher-order optimizations, demonstrate improved performance over these traditional methods, with ANADDH achieving the highest overall effectiveness. The advantage is attributed to ANADDH's adaptive optimization, which provides stable returns even in fluctuating market conditions.

In Fig.~\ref{fig-hedttm}(b), we examine the results for options with a 120-day maturity, where the benefits of Delta-Vega Hedging become more evident. Longer-maturity options, being more sensitive to volatility shifts, align well with the Delta-Vega and Gamma-Vega approaches, which account for both delta and Vega risks. The GVDH strategy leverages additional sensitivity adjustments, contributing to improved stability. However, ANADDH surpasses all other strategies, effectively adapting to both delta and Vega sensitivities in a dynamic manner, thus optimizing hedging performance as market conditions vary over the extended period.

\subsection{Impact of Initial Volatility on Hedging Performance}
\begin{figure}
	\centering
	\begin{tabular}{cccc} 
		\includegraphics[width = 0.45\linewidth]{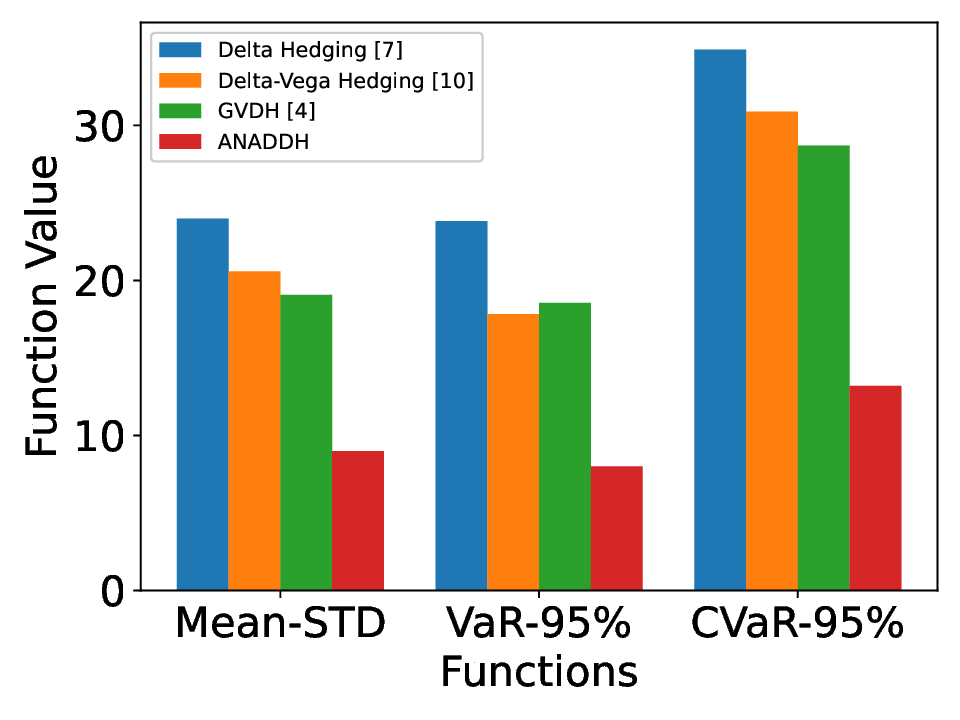} & 
		\includegraphics[width = 0.45\linewidth]{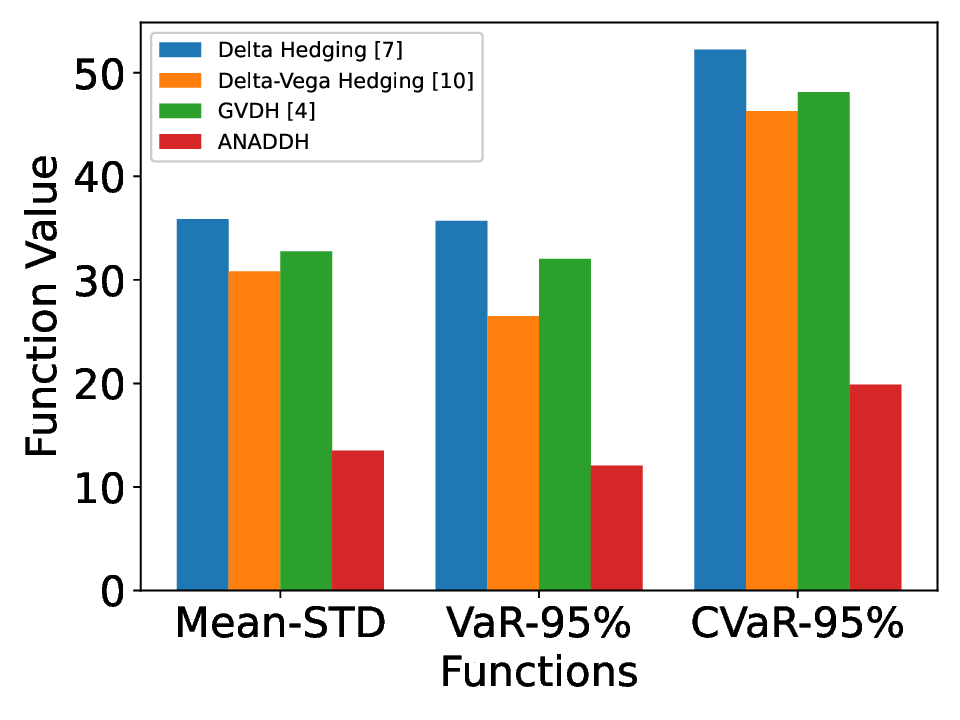} \\
		{\scriptsize (a) $\sigma = 20\%$} &
		{\scriptsize (b) $\sigma = 30\%$} \\
		\includegraphics[width = 0.45\linewidth]{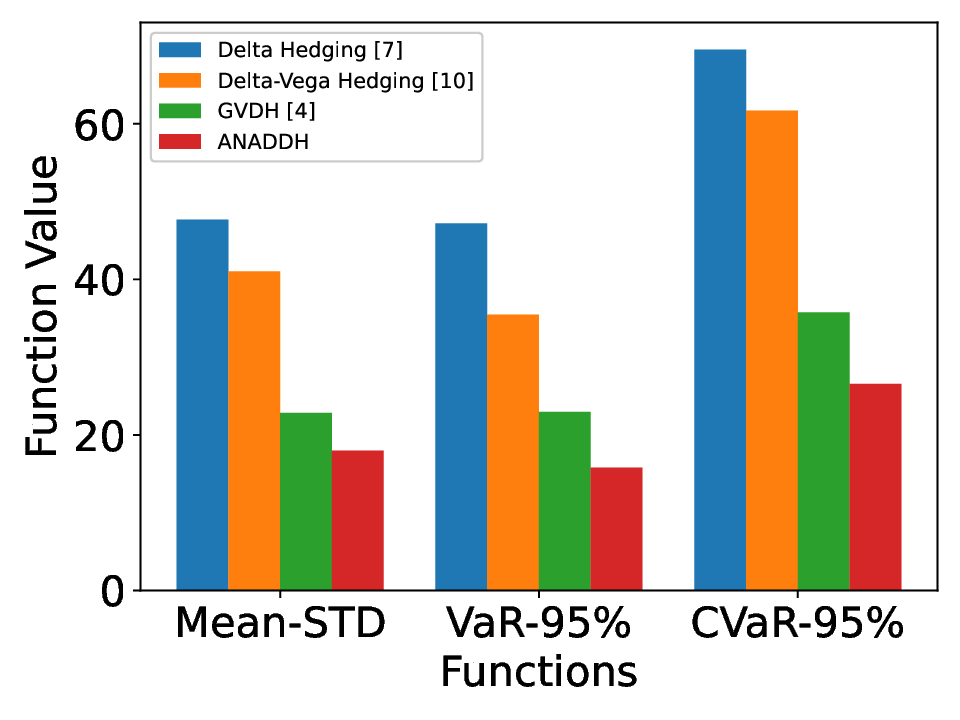} & 
		\includegraphics[width = 0.45\linewidth]{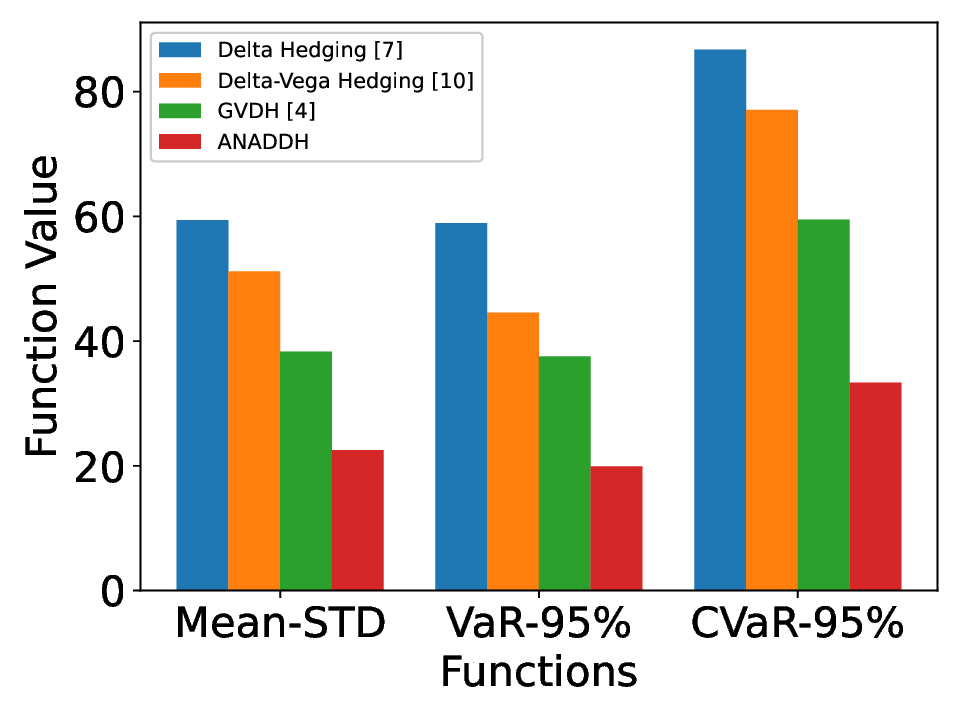} 
		\\ 
		{\scriptsize (c) $\sigma = 40\%$} &
		{\scriptsize (d) $\sigma = 50\%$}  
	\end{tabular}
	\captionsetup{font={scriptsize}}
	\caption{Performance comparison with different initial volatility $\sigma$ for the underlying asset price.}
	\label{fig-init-vol}
\end{figure}
In Fig.~\ref{fig-init-vol}, we examine the performance of each strategy under varying levels of initial volatility $\sigma = \{20\%, 30\%, 40\%, 50\%\}$, with a fixed volatility variation $\upsilon = 0.3$. We aim to validate each method's adaptability to dynamic market conditions, especially as initial volatility plays a key role in determining the sensitivity and subsequent price trajectories of options.

Initial volatility sets the stage for future price dynamics and directly influences the effectiveness of hedging actions. Lower initial volatility values generally lead to greater stability in option prices, making it easier for hedging strategies to maintain control. Conversely, higher initial volatility introduces increased uncertainty and more significant price fluctuations, posing challenges for traditional hedging approaches.

The results show that with lower initial volatility levels, such as 20\%  in Fig.~\ref{fig-init-vol}(a) and 30\% in  Fig.~\ref{fig-init-vol}(b), all strategies perform relatively well, with minimal divergence in the Mean-STD, VaR-95\%, and CVaR-95\% values, as lower volatility conditions typically involve less drastic price fluctuations, thereby presenting fewer challenges for hedging strategies. 
However, as initial volatility increases to 40\% in Fig.~\ref{fig-init-vol}(c) and 50\% in  Fig.~\ref{fig-init-vol}(d), significant differences emerge among the strategies. The traditional Delta and Delta-Vega Hedging methods show limitations in managing the heightened risk of larger price swings, as indicated by their higher Mean-STD and CVaR-95\% values. In contrast, the GVDH and ANADDH strategies, which incorporate more sophisticated risk management techniques, exhibit enhanced stability under these conditions.

ANADDH consistently outperforms all other methods across each initial volatility setting, achieving lower Mean-STD, VaR-95\%, and CVaR-95\% values. This advantage is attributed to ANADDH's adaptive Nesterov acceleration mechanism, which optimizes the response to market fluctuations by dynamically adjusting hedging actions. The flexibility of ANADDH enables it to proactively adapt its risk management strategy in response to high-volatility environments, reducing the potential for substantial losses and maintaining stable portfolio performance.

\subsection{Impact of Volatility Variation on Hedging Performance}
\begin{figure}
	\centering
	\begin{tabular}{cccc} 
		\includegraphics[width = 0.45\linewidth]{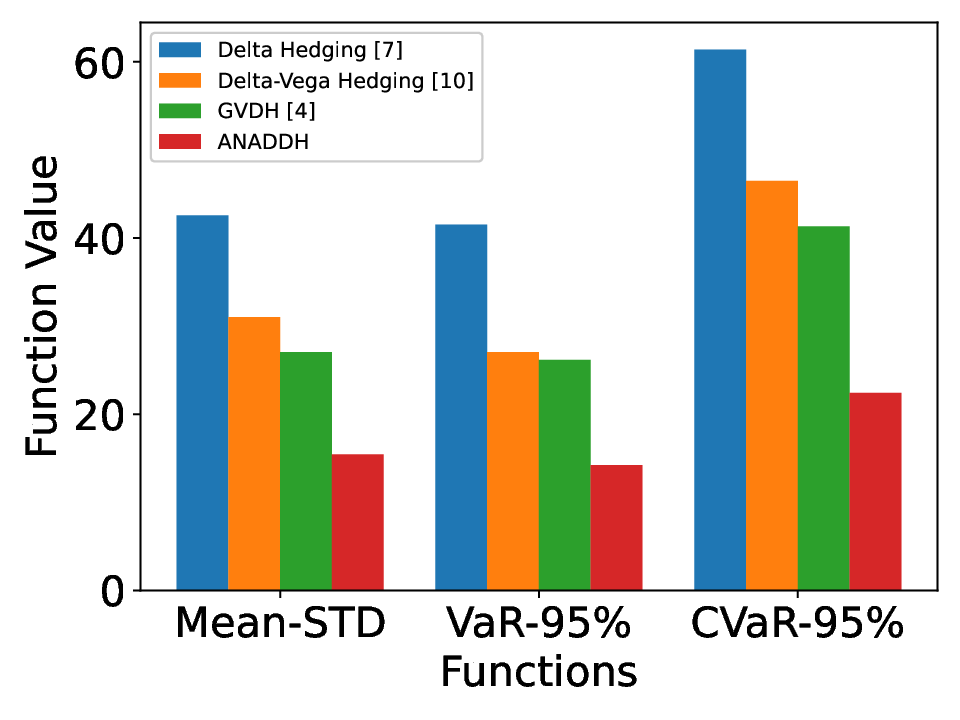} & 
		\includegraphics[width = 0.45\linewidth]{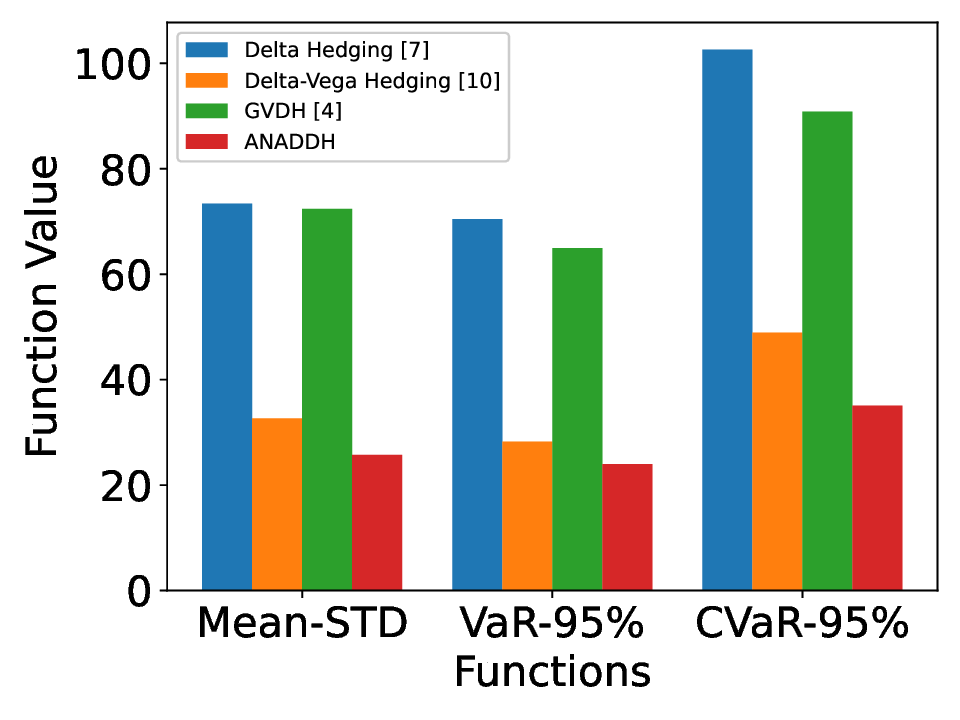} \\
		{\scriptsize (a) $\nu = 0.4$} &
		{\scriptsize (b) $\nu = 0.8$} 
	\end{tabular}
	\captionsetup{font={scriptsize}}
	\caption{Performance comparison with different  volatility of the volatility parameter.}
	\label{fig-vov}
\end{figure}
The volatility variation parameter $\upsilon$, often referred to as the volatility of volatility,  plays a critical role in determining the stability and effectiveness of hedging strategies under dynamic conditions. 
We analyze the impact of different levels of $\upsilon$, set at $0.4$ and $0.8$, on the performance of various hedging methods, as shown in Fig.~\ref{fig-vov}, which  provides insights into how rapidly fluctuating volatility impacts hedging stability and risk management, particularly for strategies that need to dynamically adjust to changing market conditions.

In Fig.~\ref{fig-vov}(a) with a relatively stable volatility environment where $\upsilon=0.4$, all hedging strategies display fairly stable performance. Traditional methods, such as Delta and Delta-Vega hedging, show moderate effectiveness, benefiting from the smoother changes in option pricing due to the lower volatility variation. However, our ANADDH framework outperforms these baseline strategies by incorporating adaptive updates that help it respond efficiently to minor fluctuations in market conditions. This results in improved stability are critical for long-term performance.

Conversely, Fig.~\ref{fig-vov}(b) illustrates the scenario with $\upsilon=0.8$, where the increased volatility of volatility introduces more frequent and pronounced fluctuations in the option's value. Under these conditions, traditional methods, such as Delta-Vega hedging, and even the state-of-the-art GVDH method encounter challenges in maintaining stability due to limited adaptability to abrupt shifts in volatility. GVDH, which finely tunes Gamma and Vega exposures, is notably impacted as these adjustments become less effective in a high dynamic environment. In contrast, ANADDH demonstrates a significant performance advantage in this scenario. Leveraging its adaptive Nesterov acceleration mechanism, ANADDH utilizes both real-time and historical gradient information to enable swift, precise updates, dynamically adapting to the rapidly evolving volatility landscape. This adaptability proves especially advantageous in high volatility markets, where proactive and responsive hedging actions are essential for managing risk and sustaining portfolio stability.

\subsection{Impact of Transaction Cost Ratios on Hedging Strategies}
\begin{figure}
	\centering
	\begin{tabular}{cccc} 
		\includegraphics[width = 0.45\linewidth]{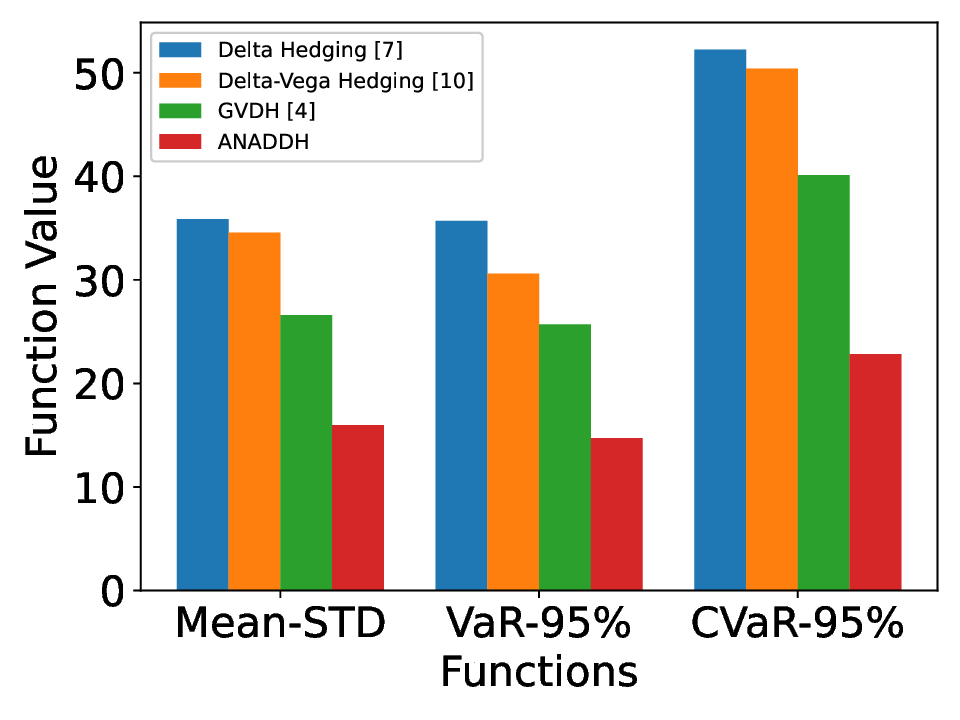} & 
		\includegraphics[width = 0.45\linewidth]{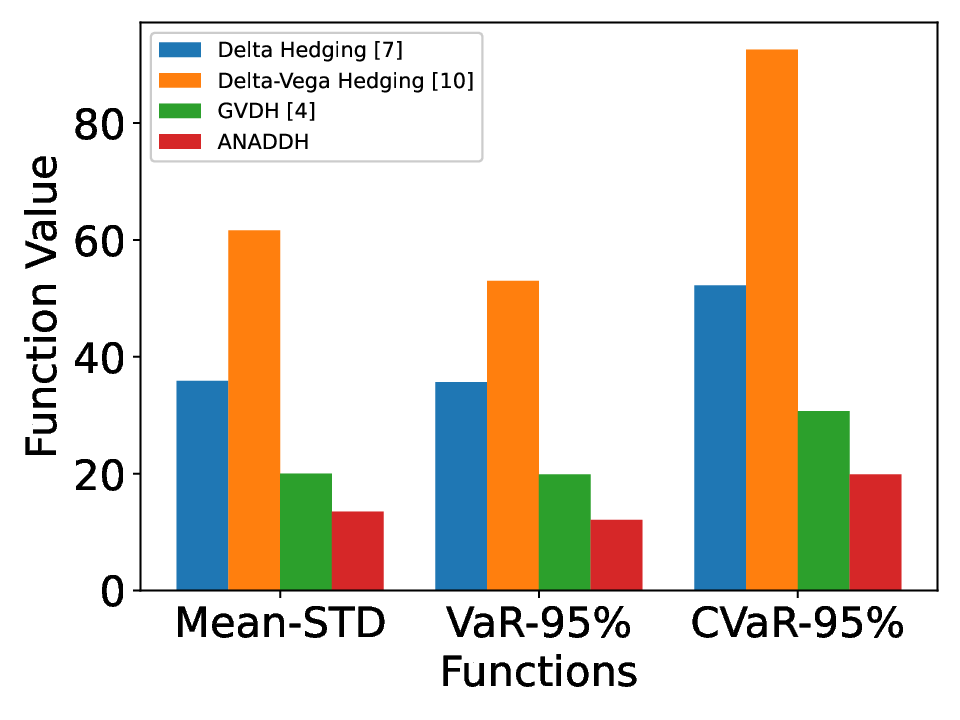} \\
		{\scriptsize (a) $0.2\%$ transaction cost} &
		{\scriptsize (b) $0.5\%$ transaction cost} \\ 
	\end{tabular}
	\captionsetup{font={scriptsize}}
	\caption{Performance comparison with increasing transaction cost ratio.}
	\label{fig-spread}
\end{figure}
The influence of transaction cost ratios on the effectiveness of hedging strategies is crucial for assessing the adaptability and cost-efficiency of risk management methods. In this analysis, we evaluate hedging performance using a 30-day ATM option as the hedging instrument, under two transaction cost scenarios, i.e., 0.2\% and 0.5\% of the option price.

In Fig.~\ref{fig-spread}(a) with a 0.2\% transaction cost ratio, Delta hedging shows strong resilience due to its minimal reliance on frequent option trades. Lower transaction costs allow this strategy to make more frequent position adjustments without prohibitive expenses, maintaining stability and closely aligning with market fluctuations. The limited cost sensitivity highlights Delta hedging's effectiveness in low-cost environments, where its direct adjustments in the underlying asset keep transaction costs manageable.
When we examine GVDH in the same context, it leverages higher-order Greeks like Gamma and Vega to fine-tune hedging actions, thereby achieving a more stable hedging performance than Delta and Delta-Vega hedging. GVDH demonstrates moderate sensitivity to transaction costs due to its more frequent adjustments to manage Gamma and Vega risk, but it remains effective in low transaction cost environments.

Conversely, in Fig.~\ref{fig-spread}(b) with a higher transaction cost ratio of 0.5\%, Delta-Vega hedging performance is notably impacted. Since Delta-Vega hedging requires more frequent adjustments to manage Vega exposure, increased transaction costs create a significant barrier. The added cost restricts the strategy's ability to adjust positions effectively, especially when volatility changes unfavorably, resulting in a less optimal approach to managing Vega sensitivity and potential drift from the desired hedging position.

Our proposed ANADDH strategy, however, demonstrates consistent robustness across both cost scenarios. It adapts dynamically by adjusting its hedging actions based on the prevailing transaction cost level, allowing it to balance stability with cost-effectiveness. ANADDH's adaptive mechanism effectively minimizes unnecessary transactions under higher cost conditions, preserving efficiency and maintaining portfolio stability. This adaptability underscores ANADDH's advantage in dynamically optimizing hedging policies under diverse transaction cost structures, ensuring reliable performance in both low and high cost environments.

\subsection{Impact of Trajectory Length on Hedging Strategy Performance}
\begin{figure}
	\centering
	\begin{tabular}{cccc} 
		\includegraphics[width = 0.45\linewidth]{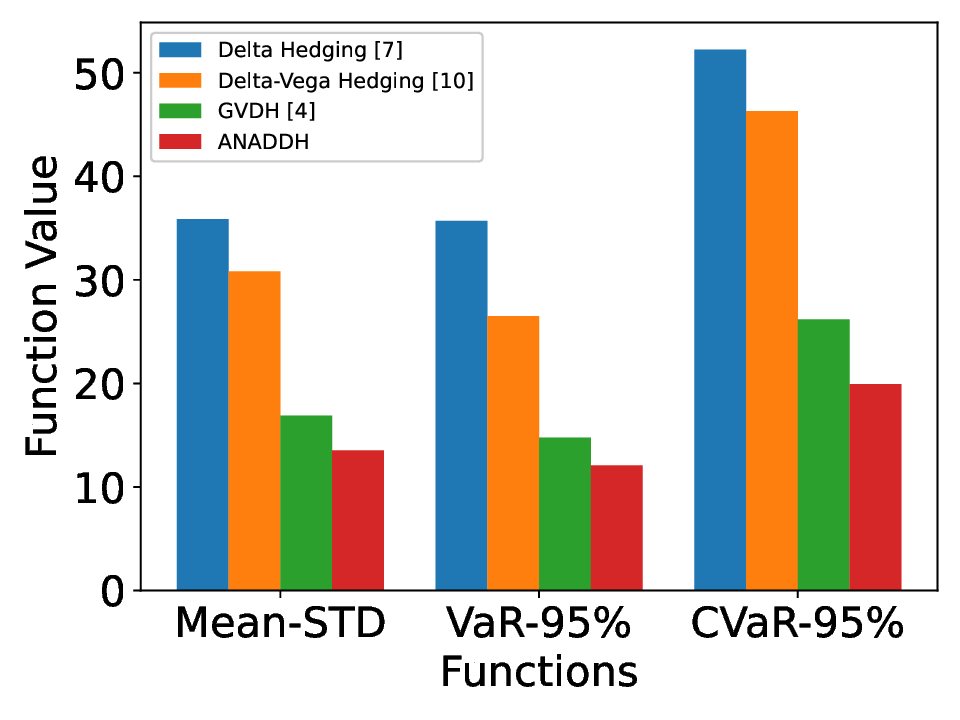} & 
		\includegraphics[width = 0.45\linewidth]{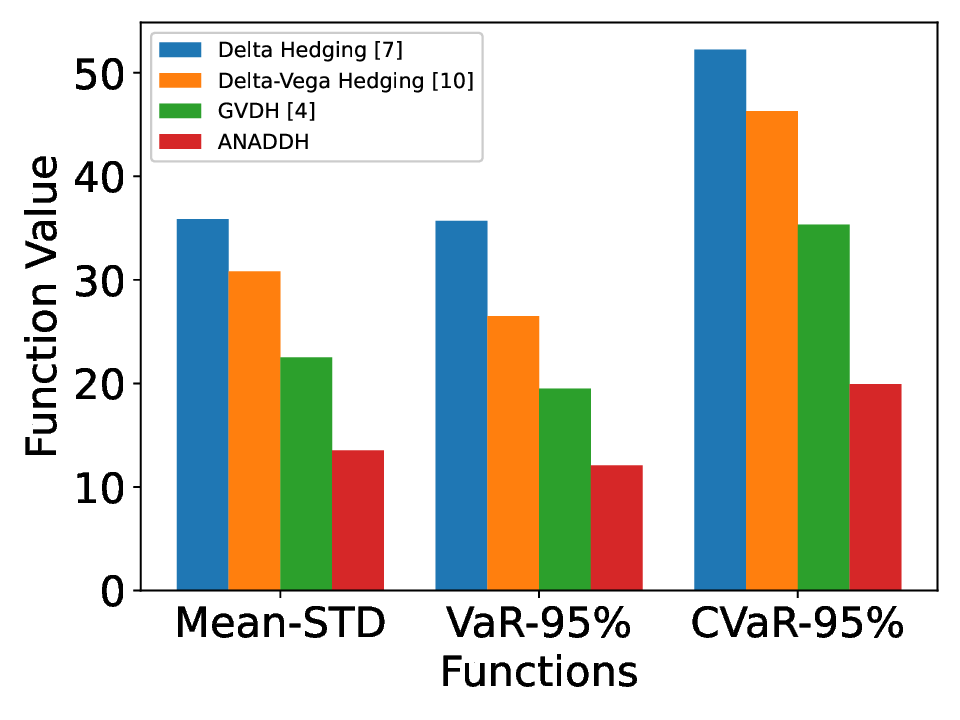} \\
		{\scriptsize (a) Trajectory Length $= 2$} &
		{\scriptsize (b) Trajectory Length $= 4$}
	\end{tabular}
	\captionsetup{font={scriptsize}}
	\caption{Performance comparison with different trajectory length.}
	\label{fig-n-step}
\end{figure}
In Fig.~\ref{fig-n-step}, we compare the performance of
our proposed ANADDH across different trajectory lengths, representing the number of steps sampled in each reinforcement learning iteration, with the other baselines.
Trajectory length determines the frequency of policy and value function updates by the actor-critic networks, impacting learning dynamics, hedging effectiveness, and computational efficiency. 
This parameter affects the learning dynamics and efficiency of the adaptive RL-based methods, i.e.,  GVDH and ANADDH, while Delta Hedging and Delta-Vega Hedging remain unaffected, as they do not employ RL.

For shorter trajectory lengths as shown in Fig.~\ref{fig-n-step}(a), frequent updates allow GVDH and ANADDH to rapidly adapt their policies based on immediate market conditions. While GVDH demonstrates responsiveness with short trajectories, our ANADDH framework consistently outperforms it, leveraging the adaptive Nesterov acceleration mechanism to make more precise and stable adjustments. This advantage highlights ANADDH's ability to maintain stability and reduce transaction costs, even when focusing on short-term market data.

In contrast, with a longer trajectory length, as shown in Fig.~\ref{fig-n-step}(b), the system integrates an extended sequence of market events before updating, fostering exploratory learning and capturing more comprehensive long-term market trends. However, as trajectory length increases, GVDH's performance declines, due to its limited ability to adapt quickly to the broader and potentially more volatile market shifts captured in longer trajectories. In comparison, our proposed ANADDH framework maintains strong performance, demonstrating its robustness and adaptability in this extended setting. 

ANADDH's adaptive Nesterov acceleration mechanism provides it with the flexibility to capitalize on broader market insights while maintaining efficient policy adjustments, effectively navigating the increased complexity of longer trajectory lengths. This capability allows ANADDH to balance immediate responsiveness with the strategic foresight necessary to optimize hedging in dynamic markets, ensuring it remains effective across both short and long term market fluctuations.

\subsection{Feasibility and Profitability of Hedging Strategies}
\begin{figure}
	\centering
	\begin{tabular}{cccc} 
		\includegraphics[width = 0.45\linewidth]{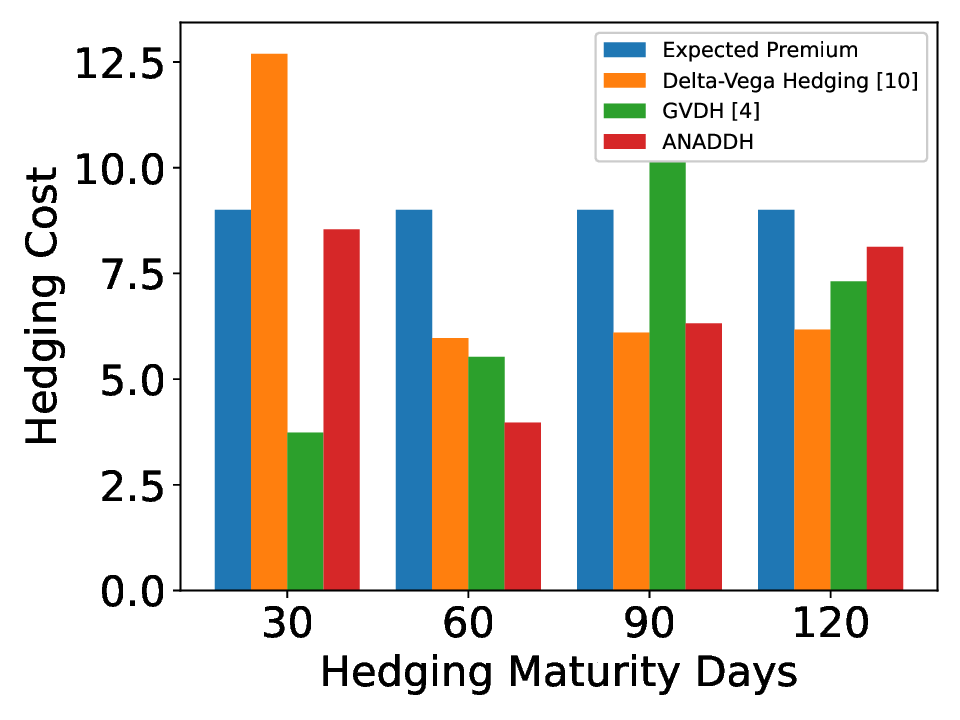} & 
		\includegraphics[width = 0.45\linewidth]{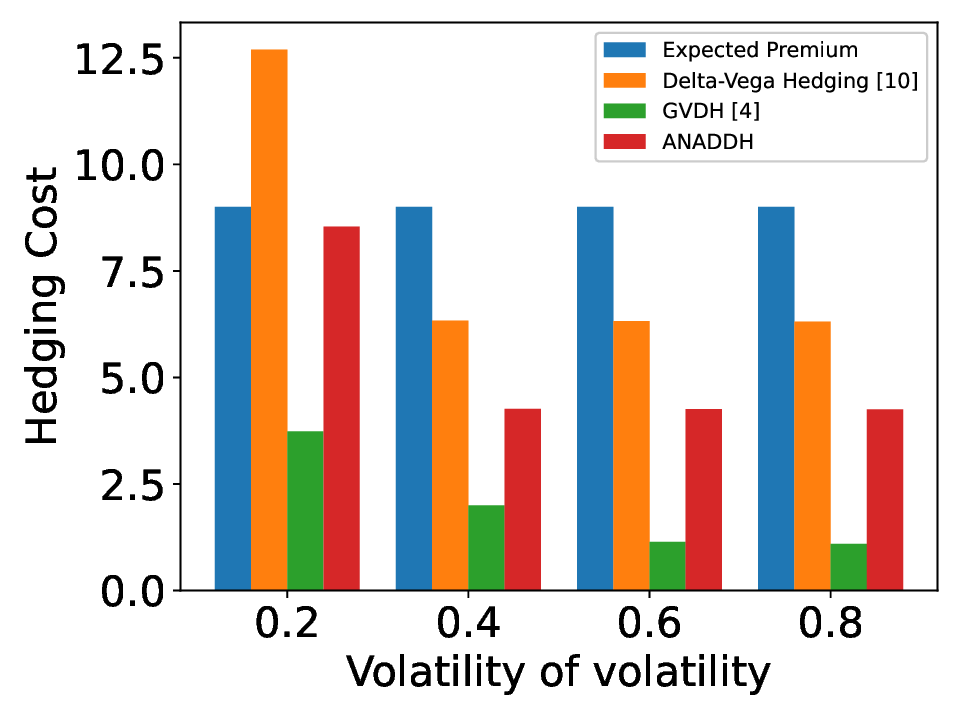} \\
		{\scriptsize (a) } &
		{\scriptsize (b)} \\
		\includegraphics[width = 0.45\linewidth]{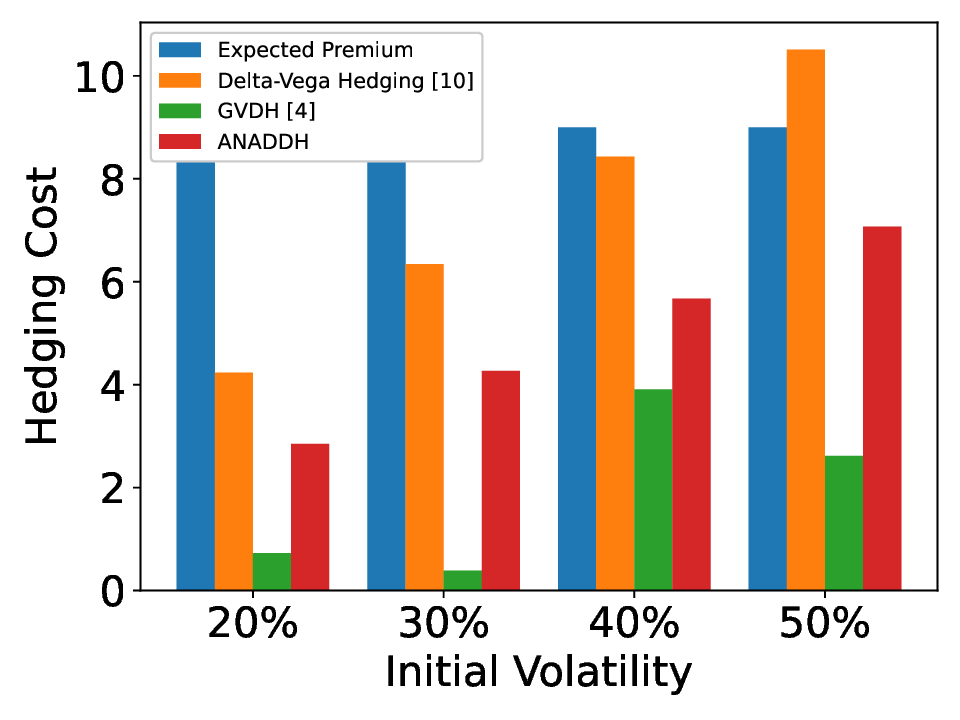} &
		\includegraphics[width = 0.45\linewidth]{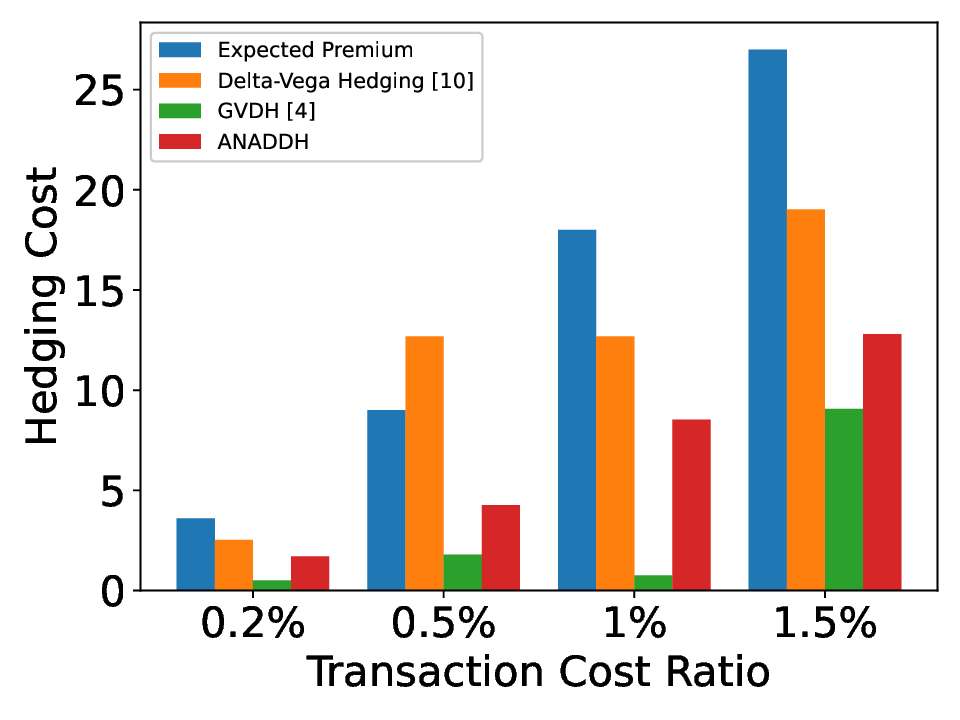} 
		\\ 
		{\scriptsize (c) } &
		{\scriptsize (d)}  
	\end{tabular}
	\captionsetup{font={scriptsize}}
	\caption{Performance comparison with the cost of hedging.}
	\label{fig-hedge-cost}
\end{figure}
To evaluate the profitability of hedging strategies, we compare the projected revenue from hedging options against the expenses incurred through hedging activities. Assuming an average arrival rate of one hedging option per day, we forecast potential profits over a 30-day period. To maintain consistency in transaction costs, we set the premium at $\kappa$ times the option price, as illustrated in Fig.~\ref{fig-hedge-cost}.

Using the Black-Scholes pricing model, we estimate each hedging option’s value to be approximately \$60. For $\kappa = 0.5\%$, the average profit generated from hedging options across various market conditions, including different option maturities, volatility variations, and initial volatility settings, amounts to about \$9, as shown in Figs.~\ref{fig-hedge-cost}(a), (b), and (c). This outcome indicates that, under typical market conditions, the revenue from hedging options comfortably exceeds the costs associated with the hedging methods, ensuring a favorable profit-to-cost balance.

We further explore a range of $\kappa$ values as 0.2\%, 0.5\%, 1\%, and 1.5\% to examine their impact on expected transaction costs and profits, depicted in Fig.~\ref{fig-hedge-cost}(d). The corresponding profits from hedging options at these values are \$3.6, \$9, \$18, and \$27, respectively. Although hedging costs rise with increasing $\kappa$ values, the rate of profit growth generally outpaces that of expenses. This positive profit-cost disparity underscores the potential for hedging option revenue to effectively cover hedging costs, enhancing the viability of the approach across different cost structures.

Not all hedging strategies, however, guarantee that hedging option revenue will fully offset hedging costs. Traditional hedging strategies may struggle to maintain profitability under certain market conditions. Although the GVDH approach sometimes yields lower hedging costs compared to our proposed ANADDH framework, this may be due to its increased trading frequency, which enhances responsiveness but also causes higher brokerage fees. This contrast highlights the importance of choosing a hedging strategy that aligns with specific cost dynamics and profitability goals, making ANADDH a flexible and cost-effective option in dynamic market environments.

\section{Conclusion}
\label{conclusion}
This paper presents an innovative approach to deep Vega hedging by integrating distributional RL with adaptive Nesterov acceleration, establishing a robust framework for managing volatility risk in financial derivatives portfolios. Leveraging distributional actor-critic networks, the proposed ANADDH enhances the adaptability of hedging strategies by enabling rapid, stable updates that respond effectively to dynamic market conditions. Our experiments show that the proposed approach significantly outperforms both traditional and the state-of-the-art deep hedging strategies, particularly in volatile environments, underscoring the advantages of adaptive optimization techniques in financial risk management. By enhancing the integration of uncertainty modeling, optimization, and risk mitigation, this work establishes a foundation for future research aimed at advancing derivatives trading strategies. Our approach highlights the potential of AI-driven innovations in financial risk management, paving the way for increasingly resilient and effective trading solutions.

\bibliographystyle{unsrtnat}
\bibliography{references}  

	
	
	
	

\newpage
\appendix
\section{Data Generation Process}
\label{A1}
\subsection{Asset Prices and Instantaneous Volatility}
Our data generation process begins with the simulation of future prices for the underlying asset using a predefined stochastic model. At each time step, asset prices are generated based on the current price, incorporating factors like expected return, volatility, and random shocks to represent potential future trajectories of the asset.

\subsection{Trading Environment Initialization}
The trading environment is initialized through the generation of random normal variables that serve as the stochastic components of the SABR (Stochastic Alpha Beta Rho) model~\cite{hagan2002managing}. We utilize standard normal random variables, denoted as $\bm{q_s}$ and $\bm{q_i}$, to inject stochastic elements into the price and volatility models, respectively. Specifically, $\bm{q_s} \sim \mathcal{N}(0,1)$ introduces randomness affecting the asset price, and $\bm{q_i} \sim \mathcal{N}(0,1)$ captures the randomness impacting the volatility. These variables are structured as matrices in $\mathbb{R}^{N \times (T+1)}$, where $N$ is the number of simulation runs and $T$ represents the number of discrete time steps within each simulation. This structured approach allows for a comprehensive exploration of price and volatility dynamics over time, providing a robust framework for analyzing the potential outcomes under various market conditions.

The proposed data generation framework is essential for simulating realistic market environments where asset prices and volatility exhibit random fluctuations. By capturing the full spectrum of possible price movements and volatility scenarios, our model ensures that the strategies developed are tested against a diverse set of conditions, enhancing the robustness and applicability of the hedging strategies derived from our research.

\subsection{Stochastic Modeling of Asset Prices and Volatility}
\label{A2}
To accurately simulate market dynamics, our data generation process intricately models the correlation between asset prices and volatility. This is achieved through the generation of correlated volatility variables $\bm{q_v}$, which are computed using the correlation coefficient $\rho$ as
\begin{equation}
	\bm{q_v} = \rho \cdot \bm{q_s} + \sqrt{1 - \rho^2} \cdot \bm{q_i},
\end{equation}
where $\bm{q_v} \in \mathbb{R}^{N \times (T+1)}$ represents the volatility component, ensuring the correct correlation structure between the asset price and its volatility.

\subsection{Initialization and Iterative Updates}
The trading environment is initialized with predefined parameters for the initial asset price and volatility. These values are then updated iteratively at each time step $t$ in the simulation, following the equations specified by the SABR model. The volatility at time $t+1$ is updated using
\begin{equation}
	\sigma_{t+1} = \sigma_t \cdot \exp\left(-\frac{\upsilon^2}{2} \Delta t + \upsilon \cdot q_v^{t,i} \cdot \sqrt{\Delta t}\right),
\end{equation}
for $t = 0, 1, \ldots, T-1$ and $i = 0, 1, \ldots, N-1$, reflecting the stochastic nature of volatility changes.

\subsection{Asset Price Updates}
Similarly, asset price updates at each time step are governed by the current asset price, the drift rate $\mu$, and the current volatility, alongside a random component influenced by $\bm{q_s}$ 
\begin{equation}
	\begin{aligned}
		P_{t+1} = P_t \cdot \exp\left((\mu - \frac{(\sigma_t \cdot P_t^{\beta - 1})^2}{2}) \Delta t + \sigma_t \cdot P_t^{\beta - 1} \cdot \sqrt{\Delta t} \cdot \bm{q_s}^{t,i}\right),
	\end{aligned}
\end{equation}
for each time step $t = 0, 1, \ldots, T-1$ and $i = 0, 1, \ldots, N-1$. This model enables the generation of multiple potential future paths for the asset's price, capturing the inherent uncertainties and dynamics of the financial market.

These detailed simulations of asset prices and volatility allow for a comprehensive exploration of financial strategies under various market conditions, providing valuable insights into the behavior of financial instruments in response to changes in market dynamics.

\begin{table}[htbp]
	\centering
	\caption{Variable Description in Environment Modeling}
	\begin{tabular}{ll}
		\hline
		\textbf{Variable} & \textbf{Description} \\
		\midrule
		$P_0$ & The current price of the underlying asset\\
		$\sigma$ & The volatility of the asset \\
		$\mu$ & Drift rate \\
		$\rho$ & Correlation coefficient between $\bm{q_s}$ and $\bm{q_i}$ \\
		$\upsilon$ & Variation of volatility \\
		$N$ & Number of simulations \\
		$T$ & Number of time periods \\
		$K$ & The strike price of the Option\\
		$F$ &The expected future price of the underlying asset at the time of maturity\\
		$t_{\text{opt}}$ & The time to maturity of options\\
		$O_c$ &  The value of call option contracts \\
		$O_p$ &  The value of put option contracts \\
		$\Delta_c$ & Delta of the call option\\
		$\Delta_p$ & Delta of the put option\\
		$V$ & Vega of both call and put options\\
		\hline
	\end{tabular}
\end{table}
\subsection{Implied Volatility of Options}
The simulated instantaneous volatilities are converted to option implied volatilities using the SABR model's formula, aligning with conventional option pricing methods. Implied volatility reflects market expectations of future volatility and is derived from the prices of actively traded options.

The forward price $F$ is computed by compounding the current spot price $P_0$ at the risk-free rate $r$, adjusted for dividends $q$ over the option's duration $t_{\text{opt}}$
\begin{equation}
	F = P_0 \times e^{(r - q) \cdot t_{\text{opt}}}.
\end{equation}
Intermediate variables $x$ and $y$ are defined to simplify the computation of implied volatility, based on the forward price $F$ and the strike price $K$
\begin{equation}
	x = (F \cdot K)^{\frac{1-\beta}{2}}, \quad y = (1 - \beta) \cdot \ln\left(\frac{F}{K}\right).
\end{equation}
Auxiliary variables $\Lambda$, $\Psi$, $\Phi$, and $\chi$ are calculated to assist in the SABR model's computations, based on these intermediate variables
\begin{equation}
	\begin{aligned}
		\Lambda &= \frac{\sigma}{x \cdot \left(1 + \frac{y^2}{24} + \frac{y^4}{1920}\right)},\\
		\Psi &= 1 + t_{\text{opt}} \cdot \left(\frac{(1 - \beta)^2 \cdot \sigma^2}{24 \cdot x^2} + \frac{\rho \cdot \beta \cdot \sigma \cdot \upsilon}{4 \cdot x} + \frac{\upsilon^2 \cdot (2 - 3 \cdot \rho^2)}{24}\right),\\
		\Phi &= \frac{\upsilon \cdot x}{\sigma} \cdot \ln\left(\frac{F}{K}\right),\\
		\chi &= \ln\left(\frac{\sqrt{1 - 2 \cdot \rho \cdot \Phi + \Phi^2} + \Phi - \rho}{1 - \rho}\right).
	\end{aligned}
\end{equation}

The implied volatility $\sigma_{\text{imp}}$ is derived from the observed market prices of options and accounts for the relationship between the strike price, forward price, and other model parameters
\begin{equation}
	\sigma_{\text{imp}} = \begin{cases} 
		\frac{\sigma \cdot \Psi}{F^{1 - \beta}}, & \text{if } F = K \\
		\Lambda \cdot \Psi \cdot \frac{\Phi}{\chi}, & \text{otherwise}
	\end{cases},
\end{equation}
which ensures comprehensive coverage of the dynamics between asset prices and volatility, providing a robust framework for accurate options pricing and risk assessment.

\subsection{Liability Portfolio Profiles}
We detail the generation and iterative updating of ATM options' prices and risk profiles, including deltas, gammas, and vegas, which are crucial for managing the liabilities associated with these option contracts in the following. 
We update these attributes step-wise for non-expired options throughout the trading simulation.

The cumulative distribution function of a standard normal distribution, denoted $\mathcal{N}(d)$, is used to evaluate the probability that a standard normal variable is less than or equal to $d$. This function is pivotal in assessing the likelihood of exercise for call options and the probability of non-exercise for put options. The variable $d$ standardizes measures such as the option's price, the underlying asset's price, time to maturity, risk-free rate, and volatility, defined as
\begin{equation}
	\begin{aligned}
		d_1 &= \frac{\ln\left(\frac{P_0}{K}\right) + \left(r - q + \frac{\sigma_{\text{imp}}^2}{2}\right)t_{\text{opt}}}{\sigma_{\text{imp}} \sqrt{t_{\text{opt}}}}, \\
		d_2 &= d_1 - \sigma_{\text{imp}} \sqrt{t_{\text{opt}}}.
	\end{aligned}
\end{equation}
Using these metrics, the value of each option contract is calculated with the Black-Scholes formula
\begin{equation}
	\begin{aligned}
		O_c &= P_0 \mathcal{N}(d_1) - K e^{-rt_{\text{opt}}} \mathcal{N}(d_2), \\
		O_p &= K e^{-rt_{\text{opt}}} \mathcal{N}(-d_2) - P_0 \mathcal{N}(-d_1).
	\end{aligned}
\end{equation}

Key to hedging strategies are the delta and vega metrics. Delta measures an option's price sensitivity to changes in the price of the underlying asset
\begin{equation}
	\begin{aligned}
		\Delta_c &= e^{-qt_{\text{opt}}} \cdot \mathcal{N}(d_1), \\
		\Delta_p &= - e^{-qt_{\text{opt}}} \cdot \mathcal{N}(-d_1),
	\end{aligned}
\end{equation}
for call and put options, respectively. Vega measures the sensitivity of the option price to changes in implied volatility
\begin{equation}
	\begin{aligned}
		V &= P_0 \sqrt{t_{\text{opt}}} e^{-\frac{d_1^2}{2}} \sqrt{\frac{1}{2\pi}}.
	\end{aligned}
\end{equation}
Daily management of Delta exposure and regular adjustments to Vega exposure are necessary to mitigate risks associated with market fluctuations in spot prices and volatility conditions. The number of options per time step is modeled using a Poisson distribution with a predetermined rate, allowing for the generation of options specifying time to maturity, strike prices, and buy or sell positions.

\section{Convergence Analysis}
\label{convergence}
First, we rewrite the gap between the current objective function value at $\bm{\theta^r}$ and the optimal function value at $\bm{\theta^*}$ by introducing the objective function value at the auxiliary point $\bm{y_r}$ as 
\begin{equation}  
	\begin{aligned}
		\mathcal{L}(\bm{\theta^r}) - \mathcal{L}(\bm{\theta^*}) = \mathcal{L}(\bm{\theta^r}) - \mathcal{L}(\bm{y_r}) + \mathcal{L}(\bm{y_r})- \mathcal{L}(\bm{\theta^*})
		.\end{aligned}
\end{equation}
According to the quadratic upper bound, we have 
\begin{equation}  
	\begin{aligned}
		\mathcal{L}(\bm{\theta^r}) - \mathcal{L}(\bm{y_r}) \leq \bm{\triangledown} \mathcal{L}(\bm{y_r})^T\bm{(\bm{\theta^r - y_r})}  + \frac{L}{2}\bm{||\theta^r - y_r||^2_2} 
		,\end{aligned}
\end{equation}
and the linear lower bound, we have 
\begin{equation}  
	\begin{aligned}
		\mathcal{L}(\bm{y_r})- \mathcal{L}(\bm{\theta^*})
		\leq  \bm{\triangledown} \mathcal{L}(\bm{y_r})^T\bm{(\bm{y_r - \theta^*})}
		,\end{aligned}
\end{equation}
which leads to the upper bound 
\begin{equation}  
	\begin{aligned}
		\mathcal{L}(\bm{\theta^r}) - \mathcal{L}(\bm{\theta^*}) 
		&\leq \bm{\triangledown} \mathcal{L}(\bm{y_r})^T\bm{(\bm{\theta^r - \theta^*})} + \frac{L}{2}\bm{||\theta^r - y_r||^2_2}
		.\end{aligned}
	\label{up-35}
\end{equation}
If we regard the upper bound in (\ref{up-35}) as a function w.r.t. $\bm{\theta^r}$, we can get the minimizer of this upper bound as
\begin{equation} 
	\begin{aligned}
		\bm{\theta^{r*}} = \bm{y_r} - \frac{1}{L}\bm{\triangledown \mathcal{L}(y_r)}
		.\end{aligned}
\end{equation}
We plug $\bm{\theta^{r*}}$ into the upper bound in (\ref{up-35}) to obtain a tighter bound as
\begin{equation}  
	\begin{aligned}
		&\mathcal{L}(\bm{\theta^r}) - \mathcal{L}(\bm{\theta^*}) \\
		&\leq  \bm{\triangledown} \mathcal{L}(\bm{y_r})^T(\bm{y_r - \theta^*} - \frac{1}{L} \bm{\triangledown} \mathcal{L}(\bm{y_r})) + \frac{L}{2}||\frac{1}{L} \bm{\triangledown} \mathcal{L}(\bm{y_r})||^2_2\\
		&= \bm{\triangledown} \mathcal{L}(\bm{y_r})^T(\bm{y_r - \theta^*)} - \frac{1}{2L}||\bm{\triangledown} \mathcal{L}(\bm{y_r})||^2_2\\
		& = L\bm{(y_r - \theta^{r*})^T}\bm{(y_r - \theta^*)} -\frac{L}{2}\bm{||y_r - \theta^{r*}||^2}
		.\end{aligned}
\end{equation}
Since $\bm{\theta^{r*}}$ is the miminizer for the right hand side of (\ref{up-35}), we have 
\begin{equation}  
	\begin{aligned}
		\mathcal{L}(\bm{\theta^r}) - \mathcal{L}(\bm{\theta^*}) 
		\leq  L\bm{(y_r - \theta^r)^T}\bm{(y_r - \theta^*)} -\frac{L}{2}\bm{||y_r - \theta^r||^2}
		.\end{aligned}
	\label{con1}
\end{equation}
Following the similar idea, by replacing $r$ by $r+1$ in  (\ref{con1}), we have 
\begin{equation}  
	\begin{aligned}
		\mathcal{L}(\bm{\theta^{r + 1}}) - \mathcal{L}(\bm{\theta^*}) \leq 
		L\bm{(y_{r+1} - \theta^{r + 1})^T(y_{r+1} - \theta^*)} - \frac{L}{2}\bm{||y_{r+1} - \theta^{r + 1}||^2_2}
		,\end{aligned}
	\label{con2}
\end{equation}
and replacing $\bm{\theta^*}$ by $\bm{\theta^r}$ in (\ref{con2}), we have
\begin{equation}  
	\begin{aligned}
		\mathcal{L}(\bm{\theta^{r + 1}}) - \mathcal{L}(\bm{\theta^r}) \leq 
		L\bm{(y_{r+1} - \theta^{r + 1})^T(y_{r+1} - \theta^r)} - \frac{L}{2}\bm{||y_{r+1} - \theta^{r + 1}||^2_2}.
	\end{aligned}
	\label{con3}
\end{equation}
Combining (\ref{con2}) and (\ref{con3}), we have
\begin{equation} 
	\begin{aligned}
		\mathcal{L}(\bm{\theta^r}) - \mathcal{L}(\bm{\theta^*}) \leq  
		L\bm{(y_{r+1} - \theta^{r + 1})^T(\theta^r - \theta^*)}
		.\end{aligned}
	\label{con4}
\end{equation}
If we define $\delta_r = \mathcal{L}(\bm{\theta^r}) - \mathcal{L}(\bm{\theta^*})$ and $\delta_{r+1} = \mathcal{L}(\bm{\theta^{r + 1}}) - \mathcal{L}(\bm{\theta^*})$, then by combining (\ref{con2}) and (\ref{con4}), we have that
\begin{equation} 
	\begin{aligned}
		t_r\delta_{r+1} + (1 - t_r)\delta_r 
		\leq 
		L\bm{(y_{r+1} - \theta^{r + 1})}^T(t_r\bm{y_{r+1}} + (1 - t_r)\bm{\theta^r - \theta^*}) 
		- \frac{L}{2}t_r\bm{||y_{r+1} - \theta^{r + 1}||^2_2}.
	\end{aligned}
	\label{eq42}
\end{equation}
Based on the definition of the updating rule of parameter $t_r$, we get 
\begin{equation}  
	\begin{aligned}
		t_{r+1} - t_{r+1}^2 = - t_r^2 
		,\end{aligned}
\end{equation}
from which we can get
\begin{equation} 
	\begin{aligned}
		t_{r+1}[t_{r+1}\delta_{r+1} + (1 - t_{r+1})\delta_r] =  t_{r+1}^2\delta_{r+1} + (t_{r+1} - t_{r+1}^2)\delta_r = t_{r+1}^2\delta_{r+1}  - t_{k}^2\delta_r
		.\end{aligned}
	\label{t-eq}
\end{equation}
For notation simplicity, we define 
\begin{equation}  
	\begin{aligned}
		\bm{\theta} = t_{r+1}\bm{y_{r+1}} + (1 - t_{r+1})\bm{\theta^r - \theta^*},
	\end{aligned}
\end{equation}
by combining (\ref{eq42}) and (\ref{t-eq}), we have 
\begin{equation} 
	\begin{aligned}
		t_{r+1}^2\delta_{r+1}  - t_{k}^2\delta_r \leq 
		t_{r+1}L\bm{(y_{r+1} - \theta^{r + 1})^T\theta} - \frac{L}{2}||t_{r+1}\bm{(y_{r+1} - \theta^{r + 1})}||^2_2.
	\end{aligned}
	\label{eq46}
\end{equation}
According to the fact that
\begin{equation} 
	\begin{aligned}
		2\bm{a^Tb - ||a||^2 = ||b||^2 - ||b - a||^2}
		,\end{aligned}
\end{equation}
(\ref{eq46}) can be rewritten as
\begin{equation} 
	\begin{aligned}
		t_{r+1}^2\delta_{r+1}  - t_{k}^2\delta_r \leq  \frac{L}{2}\bm{||\theta||^2} -\frac{L}{2} ||t_{r+1}\bm{\theta^{r + 1}} + (1 - t_{r+1})\bm{\theta^r - \theta^*}||^2
		.\end{aligned}
	\label{eq-48}
\end{equation}
From the definition of the updating rule of $\bm{y_{r}}$, we have
\begin{equation} 
	\begin{aligned}
		t_{r+1} \bm{y_{r+1}} + (1 - t_{r+1})\bm{\theta^r} = t_r\bm{\theta^r} + (1 - t_r)\bm{\theta^r}
		,\end{aligned}
	\label{eq-49}
\end{equation}
then, we define
\begin{equation} 
	\begin{aligned}
		\bm{u_r} = t_{r+1}\bm{y_{r+1}} + (1 - t_{r+1})\bm{\theta^r} - \bm{\theta^*}
		,\end{aligned}
	\label{eq-50}
\end{equation}
by combining (\ref{eq-48})-(\ref{eq-50}), we have 
\begin{equation} 
	\begin{aligned}
		t_{r+1}^2\delta_{r+1}  - t_{k}^2\delta_r  \leq 
		\frac{L}{2}(\bm{||u_r||^2_2 - ||u_{r+1}||^2_2})
	\end{aligned}
\end{equation}
Summing these inequalities from $0$ to $r -1$, we obtain
\begin{equation} 
	\begin{aligned}
		t_r^2\delta_r - t_0^2\delta_0  \leq \frac{L}{2} \left(\bm{||u_0||^2_2 - ||u_r||^2_2}\right) \leq \frac{L}{2} \bm{||u_0||^2_2}
		.\end{aligned}
\end{equation}
Since 
\begin{equation} 
	\begin{aligned}
		\bm{u_{0}}  = t_{1}\bm{y_{1}} + (1 - t_{1})\bm{\theta^0} - \bm{\theta^*}  = \bm{\theta^0} - \bm{\theta^*}
		,\end{aligned}
\end{equation}
and $t_0 = 0$ still follows the designed updating rule and $t_r \geq \frac{r}{2}$ can be proved by induction. 
We know $t_1 = 1 \geq \frac{1 + 1}{2}$, and assuming $t_r \geq \frac{r+1}{2}$, we can show that 
\begin{equation} 
	\begin{aligned}
		(2t_{r+1} - 1)^2 = 1 + 4 t_r^2 \geq  1 + (r+1)^2 \geq (r+1)^2,
	\end{aligned}
\end{equation}
which leads to $t_{r+1} \geq \frac{r+2}{2}$.
Thus, we can get the convergence performance as 
\begin{equation}
	\begin{aligned}
		\mathcal{L}(\bm{\theta^r}) - \mathcal{L}(\bm{\theta^*})  \leq 
		\frac{2L\bm{||\theta^0 - \theta^*||^2_2}}{(r +1)^2}.
	\end{aligned}
	\label{NAG-C1}
\end{equation}

\end{document}